\documentclass{scrarticle}

\usepackage[english]{babel}

\usepackage[letterpaper,top=2cm,bottom=2cm,left=3cm,right=3cm,marginparwidth=1.75cm]{geometry}

\usepackage{amsmath}
\usepackage{graphicx}
\usepackage[colorlinks=true, allcolors=blue]{hyperref}
\usepackage{natbib}
\usepackage{caption}
\usepackage{subcaption}
\usepackage{color}
\usepackage{lineno}
\usepackage{graphicx}
\usepackage{amsmath,amsfonts,amssymb,commath,adjustbox}
\newcommand{\bftab}{\fontseries{b}\selectfont}

\usepackage[normalem]{ulem}
\useunder{\uline}{\ul}{}

\title{How direct is the link between words and images?\thanks{Accepted in the Mental Lexicon Journal: \url{https://benjamins.com/catalog/ml} }}
\renewcommand\footnotemark{}
 
\author{Hassan Shahmohammadi, Maria Heitmeier, Elnaz Shafaei-Bajestan,\\ Hendrik P. A. Lensch, and R. Harald Baayen\\\\
University of Tübingen\\ \texttt{hassan.shahmohammadi@uni-tuebingen.de}}

\date{}

\begin{document}





\maketitle

\begin{abstract}
\noindent
Current computational models capturing words' meanings mostly rely on textual corpora. While these approaches have been successful over the last decades, their lack of grounding in the real world is still an ongoing challenge. In this line of work, \citet{Gunther2020Images} proposed a behavioral experiment to investigate the relationship between words and embodied experience. In their setup, participants were presented with a target noun and a pair of images, one chosen by their computational model and another chosen randomly. The participants were then asked to select the image that best matched the target noun. The results showed that, in the majority of cases, participants preferred the image selected by the model above chance level. \citet{Gunther2020Images} interpret this result as supporting the possibility of  
a direct link between words and embodied experience. We took their experiment as point of departure and addressed the following research questions in the context of visually grounded word representations. 1. Apart from utilizing visually embodied simulation of given images, what other strategies might subjects have used to solve this task? To what extent does this setup rely on visual information from images? Can it be solved using purely textual representations? { 2.}
Do current visually grounded embeddings explain subjects' selection behavior better than textual embeddings?
{ 3. Does visual grounding improve the semantic representations of both concrete and abstract words? } To address these questions, we designed novel experiments by using pre-trained textual and visually grounded word embeddings. Our experiments reveal that subjects' selection behavior is explained to a large extent based on purely text-based embeddings and word-based similarities, suggesting a minor involvement of active embodied experiences.  Visually grounded embeddings offered modest advantages over textual embeddings only in certain cases. These findings indicate that the experiment by \citet{Gunther2020Images} may not be well suited for tapping into the perceptual experience of participants, and therefore the extent to which it measures visually grounded knowledge is unclear. 
\end{abstract}

\newpage

\section{Introduction}

How do knowledge and specifically the knowledge of word meanings emerge? For more than 2000 years, it was thought that mental representations of meanings at least resemble the perceptual states producing them \citep{Barsalou1999PerceptualSystems, Zwaan2005EmbodiedComprehension}. Though heavily debated in the early to mid 20th century \citep[overview in][]{Barsalou1999PerceptualSystems}, grounded cognition has regained acceptance over the last decades \citep{barsalou2010grounded} and is especially prominent in cognitive linguistics \citep[e.g.][]{Lakoff1987WomenThings, Langacker1999ALinguistics}. It is supported by a range of empirical studies reporting evidence in favour of an embodied view of cognition \citep[e.g.][]{Simmons2005PicturesReward, Martin2007TheBrain, Goldstone1995EffectsPerception, Solomon2001RepresentingLocally, Solomon2004PerceptualVerification, Barsalou2008GroundedCognition}. 

Nevertheless, over the last 30 years, computational models of meaning called Distributed Semantic Models (DSM) have emerged, which derive meaning representations entirely from text corpora, i.e., from language alone. They are based on the hypothesis that words occurring in the same context are semantically related \citep{Harris1954DistributionalStructure}. Prominent examples are Latent Semantic Analysis \citep[LSA;][]{Landauer1997AKnowledge.}, word2vec \citep{mikolov2013efficient} and GloVe \citep{Pennington2014Glove:Representation}. DSM have been applied widely and successfully not only in Natural Language Processing \citep[NLP; e.g.][]{wang2019evaluating}, but also in cognitive science \citep[overview in][]{Gunther2019}. For example, they have been shown to be able to predict aspects of brain imaging data such as fMRI and EEG \citep{Bulat2017SpeakingBrain, Hollenstein2019CogniVal:Evaluation, anderson2015reading}, explain variance in a range of behavioural data \citep{Mandera2017ExplainingValidation, Westbury2014YouJudgments, Westbury2019WrigglyFunny, marelli2018database}, and even reflect relations between colours and geographical locations respectively \citep{Abdou2021CanColor, louwerse2009greographical}.

All these purely text-based methods, while providing a practical way for dealing with large-scale data, suffer from an obvious limitation often referred to as the symbol grounding problem \citep{harnad1990symbol}. That is, the meanings of words are solely based on other words without links to the outside world. Moreover, they also have practical problems; for example, because of the fundamental assumption of DSM, antonyms belonging to the same topical class (e.g., {\em small} and {\em big}) typically end up very close together in purely text-based vector spaces \citep[see, e.g.,][and references cited there]{shahmohammadi2021learning}. As a consequence,  applications for, e.g., sentiment analysis, cannot well distinguish between ``it was a good movie'' and ``it was a bad movie''.\footnote{
{For recent progress for sentential negation, see \citet{anschutz2023correct}.
}}

These issues, as well as the empirical findings regarding embodied views of cognition, have sparked interest in grounding word meaning in experience. One approach to gain insight into speakers' conceptual knowledge and to obtain such representations is to simply ask participants what they associate with a number of concepts. \citet{McRae2005SemanticThings} and \citet{Buchanan2019EnglishConcepts}, among others, asked participants to provide perceptual, functional, and encyclopedic features for a range of target words, resulting in so-called feature norms. \citet{Lynott2020Lancaster} targeted the sensorimotor experience with words more specifically by asking participants to rate words on 11 sensorimotor dimensions, including auditory, visual, and interoceptive experience, but also the extent to which they involve body parts such as arms, legs, or heads. Both of these norms have been useful in cognitive science and used to, e.g., capture the semantic richness of concepts in memory \citep[e.g.][]{Grondin2009SharedConcepts, Buchanan2019EnglishConcepts}. Nevertheless, such norms do not necessarily mirror mental representations exactly, as only especially salient features may be recalled or only the features of a particular memory might be elicited \citep[e.g.][]{Cree2003AnalyzingNouns., Barsalou2003AbstractionSystems, Buchanan2019EnglishConcepts}.

A second approach has been to utilise image data to integrate visual information into word embeddings. Visual information can be extracted from images by means of image classification models. From these models, features can be extracted representing a condensed, distributed representation of an image. \citet{baroni2016grounding} differentiates between two methods of grounding textual information in vision: \textit{multimodal fusion} and \textit{cross-modal mapping}. 

\textit{Multimodal fusion} methods range from simple concatenation \citep{kiela-bottou-2014-learning, rotaru2020constructing, utsumi2022test}, to projecting both into a shared space \citep{silberer-lapata-2014-learning,hasegawa-etal-2017-incorporating, kiela-etal-2018-learning,chrupala-etal-2015-learning}, or aligning textual embeddings with image information \citep{shahmohammadi2021learning, bordes-etal-2019-incorporating}. Multimodal fusion approaches are successful in NLP \citep{bruni2014multimodal,shahmohammadi2021learning, bordes-etal-2019-incorporating}, as well as showing some promise when it comes to predicting cognitive phenomena \citep[e.g.][]{Bulat2017SpeakingBrain, Lazaridou2016MultimodalInput, Lazaridou2017MultimodalText, DeDeyne2021VisualMind, anderson2015reading, rotaru2020constructing, utsumi2022test}. However, there remain a number of open questions, such as how much image information should be allowed to be fused into textual embeddings, whether image information should be linked to individual words or words in context, and how much grounding helps when a lot of textual training data is available. These questions have been explored extensively in \citet[][henceforth \citeauthor{shahmohammadi2022language}]{shahmohammadi2022language}. They propose a new method for grounding textual embeddings in vision. This method, a multimodal fusion model \citep{baroni2016grounding}, is trained on full image descriptions instead of single words. Hence, it leverages both the textual context and all visual information. After the training phase, the algorithm can be used to obtain a visually grounded representation of any single target word even if it has not been seen {during training (so-called ``zero-shot''). Accordingly, we will refer to this model as ``ZSG'', Zero-shot Grounded, for the remainder of this paper.} In \citeauthor{shahmohammadi2022language}, ZSG was evaluated on a range of different tasks.  This allowed the authors to investigate several questions related to how much visual information should be allowed to fuse into textual embeddings, how much training on full sentences instead of single words benefits the grounding process, and finally, how the amount of textual training data available for a particular task influences the benefit of visual grounding.

\textit{Cross-modal mapping}, on the other hand, involves mapping between visual and textual information, for instance, to model how visual concepts can be translated into language. One example of such a model is proposed { in a study by G{\"{u}}nther, Petilli,  Vergallito, and Marelli (2020), \nocite{Gunther2020Images}  henceforth GPVM,}  who employ a simple linear mapping to map from textual to visual information. Their model is designed to account for the grounding of both concrete and abstract nouns.
They trained a linear mapping to map from language-based textual embeddings to visual embeddings of concrete nouns. Then, they mapped the text embeddings of a set of target nouns, including unseen concrete and abstract nouns, to the vision domain. They selected the image that was most similar to the generated image vector as the image that corresponds to the target noun. Thus they could predict images also and critically for unseen abstract nouns. In the following experiments, the images provided by this setup were compared with random control images by asking participants to select the image which best represented the target noun. They concluded that their model was able to account for the visual grounding of abstract words but that concreteness was nevertheless a significant predictor of model performance, i.e. the model performed much better on concrete than on abstract words. 

{GPVM's} model and subsequent behavioural experiments raise a number of questions and issues of how textual representations should be grounded in vision, which deserve further investigation. The model shows how knowledge based on language can be translated into knowledge based on vision. However, previous work suggests that rather than mapping from one modality to the other, combining the two sources of information into a multimodal representation might be preferable
. Notably, \citeauthor{shahmohammadi2022language} conducted various experiments comparing cross-modal mapping accounts such as the one by GPVM 
to more sophisticated ones of multimodal fusion and found that a simple linear mapping from language to vision (which they call a ``Word-Level'' model) results in representations which are much worse at predicting human relatedness and similarity judgments compared to a number of different multimodal fusion architectures (compared to ZSG, their best model, the Word-Level model was on average less accurate by 30 percentage points). In other words, the vision representations predicted from textual embeddings in a model similar to the one in {GPVM} 
do not perform well in tasks assessing words' semantic similarity and relatedness \citep[as perceived by human raters, such as the MEN task,][]{bruni2014multimodal}.

Nevertheless, the behavioural data collected by {GPVM} 
provides insights into speakers' semantic representations of words, specifically highlighting the role of visual information available in their experiments. In the present study, we therefore take the behavioural experiment by GPVM 
as point of departure and conduct novel { computational} experiments to evaluate the multimodal embeddings by \citeauthor{shahmohammadi2022language} compared to a baseline of simple textual embeddings. This is especially worthwhile since both the model underlying the behavioural experiment of GPVM 
and { the ZSG model} claim to be able to ground unseen abstract words into vision. Using current multi-modal embeddings to model participants' behavior provides useful insights for analyzing the interplay between vision and language in human cognition. 

In contrast to GPVM, 
who used their behavioural experiment as a simple verification of their model, we want to model participants' responses. As a first step, we need a task analysis, working out what participants might actually be doing in the experiment.  GPVM 
explicitly state that their model is not a processing model, i.e., they do not describe how and whether participants in their experiment actually access visual information. For instance,  it is unclear whether GPVM 
infer from their experimental results that abstract words evoke images in the mind, which subsequently might influence lexical processing. However, their mapping model appears to suggest that participants translate language representations to visual representations in order to solve the task of which presented image better matches the presented target word.

A number of both behavioural and neuropsychological studies has explored to what extent humans actually generate mental images in knowledge retrieval. It was found, for example, that perceptual characteristics (e.g., size) best accounted for verification times and errors. 
This has been taken to indicate that participants simulated concepts visually to verify their properties \citep{Solomon2004PerceptualVerification}. Further evidence comes from lesioning studies: it was found that damage to certain areas in the brain increased the probability of losing categories relying on that area. For example, if visual areas are damaged, patients are more likely to lose the animal category since it relies on visual processing \citep[overview in][]{Barsalou2008GroundedCognition}. However, this is rarely interpreted as providing evidence for concrete images being created ``inside the brain''. ``[T]he claim is not that there are pictures in the mind. Rather, the claim is that traces of visual and other experiences are (partly) reactivated and recombined in novel ways by associated words'' \citep[p. 242]{Zwaan2005EmbodiedComprehension}. Or, in the words of \citet[p. 582]{Barsalou1999PerceptualSystems}, ``[perceptual symbols] are records of the neural states that underlie  perception''. And indeed, \citet{Solomon2004PerceptualVerification} found that participants only used perceptual simulation in a feature verification task if the task could not be solved based on association alone (but see \citet[][p. 267]{Barsalou2008LanguageProcessing}: in a task where participants were presented with words for 5 seconds, and only then asked whether the word applied to an image that was next presented, involvement of the simulation system was found).

In the present paper, therefore, we aim to answer three questions:

\begin{enumerate}
    \item How can we explain the data of { GPVM} without assuming that participants generate mental images? 

    Specifically, we propose that rather than actually translating textual representations into visual representations, as suggested by the model of GPVM, 
    participants might identify the objects in the images and associate the object names with the target word. The image with closer associations to the target word would then be selected as the image that better fits the target word. { In GPVM, 
    similarity judgments appear to take place at the level of visual similarity only. We on the other hand propose that rather than at the visual level, similarity judgments occur within a semantic space influenced by our understanding of words through their co-occurrence and potential visual attributes.} For example, in one of the examples by GPVM 
    (see Table~\ref{tab:gunther_images}), the model prediction shows an image with three children, while the other random image depicts a green plant. The target word is ``childhood''. Based on the objects ``children'' and ``plant'' it can be predicted that participants  will choose the image depicting ``children'', as it is clearly more associated with ``childhood'' than ``plant''. Solving the task at hand { can therefore in principle be} accomplished without ever generating an internal image of ``childhood''.
    
    \item As discussed above, previous research nevertheless indicates that word embeddings grounded in vision are better at predicting behavioural data than purely textual embeddings \citep[e.g.][]{bruni2014multimodal, shahmohammadi2022language}. Our understanding of the task still warrants participants to make comparisons  between words {(albeit in semantic rather than in visual space}), and here, it will be interesting to see whether their responses are best predicted by purely textual \citep[GloVe/Word2Vec;][]{Pennington2014Glove:Representation, mikolov2013efficient} or by the multimodal embeddings of \citet[][]{shahmohammadi2022language}.
    
    Thus, assuming that the alternative explanation proposed above finds support, 
    a central question remains: Is participants' behaviour best accounted for by purely textual or multimodal word embeddings?
    
    \item The last question relates to the treatment of abstract words in the process of visual grounding. Abstract words are differentiated from concrete words by the (un)availability of their denotations to the human senses. Concepts become gradually more abstract as they are separated further from sensible physical entities and become more associated with mental states \citep{Barsalou2003AbstractionSystems}. Evidently, obtaining images for an abstract word such as \textit{malice} is much more difficult, if at all possible, than for a straightforwardly concrete word such as \textit{apple}. Accordingly, image corpora usually provide images exclusively for common concrete objects. Nonetheless, some abstract concepts can be elicited from the contextual situations in which these objects occur.  
    
    In the grounding approach by  \citet{shahmohammadi2022language}, initial grounding takes place for concrete words, after which other words, including abstract words, can also be grounded using zero-shot learning. They show that zero-shot learning of visual grounding is highly effective, not only for concrete words but also for abstract words. In this study, we ask: 1) Does this finding replicate using the data of GPVM? 
    If so, 2) Does the indirect grounding of abstract words afford a better understanding of the experimental results reported by GPVM? 
    
    These results are particularly interesting given the extensive evidence from case reports and behavioral and neural studies suggesting that abstract and concrete words are processed differently, involving overlapping but distinct brain regions \citep[see][for reviews]{Montefinese2019, Mkrtychian2019}. 
    
       
\end{enumerate}

The rest of the paper is structured as follows: in Section~\ref{sec:method} we first introduce in detail the model and experiment presented by GPVM, 
followed by a description of the grounding model of \citeauthor{shahmohammadi2022language}. Subsequently, in Section~\ref{sec:results} we model the experiment utilizing both textual and grounded word embeddings, aiming to answer the three questions formulated above. In Section~\ref{sec:discussion} we discuss our results.

\section{Methodology}\label{sec:method}
This section details the model and the behavioral experiment by GPVM and 
and briefly explains the \textit{ZSG} grounding approach proposed by \citeauthor{shahmohammadi2022language}.

\begin{figure}
     \centering
     \begin{subfigure}[b]{0.65\textwidth}
         \centering
         \includegraphics[width=\textwidth]{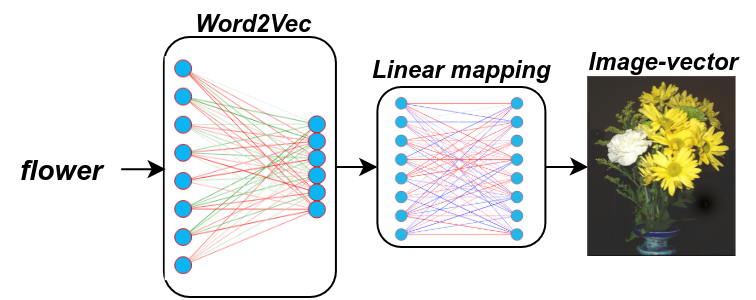}
         \caption{Cross-modal mapping model by \citet{Gunther2020Images}}
         \label{fig:marco_model}
     \end{subfigure}
     \hfill
     \begin{subfigure}[b]{\textwidth}
         \centering
         \includegraphics[width=.9\textwidth]{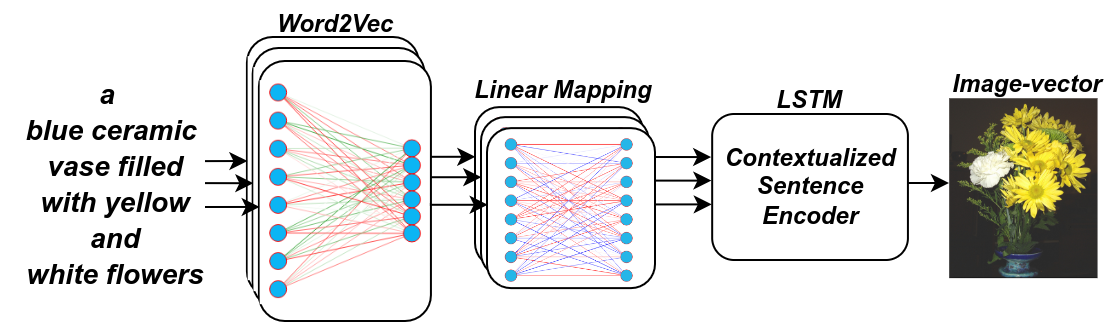}
         \caption{Multimodal fusion model by \citet{shahmohammadi2022language}}
         \label{fig:our_model}
     \end{subfigure}
     \caption{Comparison of the cross-modal mapping model by \citet{Gunther2020Images} 
     and the multimodal fusion model proposed by \citet{shahmohammadi2022language}. 
     {The latter model} takes the context into account and first applies the mapping matrix $\mathbf{M}$ to the textual embeddings of all words in the context. Then, it applies a contextualized sentence encoder to predict the image vector.}
     \label{fig:comp}
\end{figure}

\subsection{Materials from \citet{Gunther2020Images}}\label{sec:gunther}

GPVM proposed a grounding model that  
combines vision and language information. It maps  textual representations \citep[obtained from a pre-trained \textit{Word2Vec-cbow} model,][]{mikolov2013efficient} onto image vectors \citep[obtained from VGG-F, a pre-trained image classification model,][]{Chatfield2014ReturnNets} via a single linear mapping (see Figure~\ref{fig:marco_model}). It is first trained on a set of isolated words for which images are available in ImageNet \citep{deng2009imagenet} and later tested on both concrete and abstract words which did not occur in the training set. For instance, it is first trained to predict an image vector of a \textit{dog}, given a word vector of \textit{dog}. The trained model is then used to generate visual representations for unseen words including abstract words such as \textit{jealousy} or \textit{childhood}. 

GPVM 
trained two versions of their model: a prototype model, where for each word, the image representations of 100 to 200 images \citep[depending on how many were available in ImageNet,][]{deng2009imagenet} were averaged to obtain a ``prototype'' representation, and an exemplar model, for which the model was trained on 20 different image representations per word.

\begin{figure}[ht!]
     \centering
     \begin{subfigure}[b]{\textwidth}
         \centering
         \includegraphics[width=\textwidth]{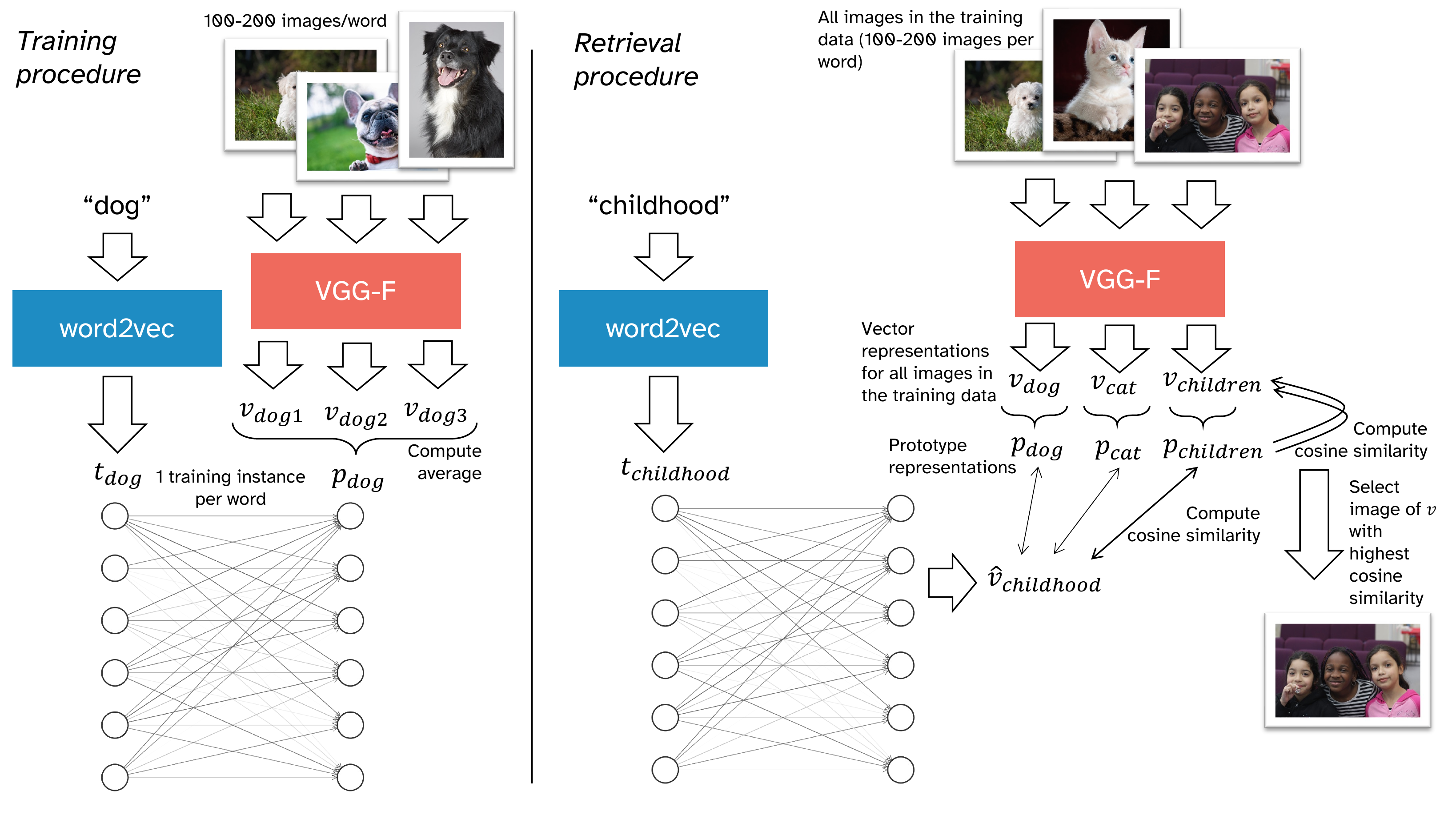}
         \caption{Prototype model}
         \label{fig:model_proto}
     \end{subfigure}
     \hfill
     \begin{subfigure}[b]{\textwidth}
         \centering
        \includegraphics[width=\textwidth]{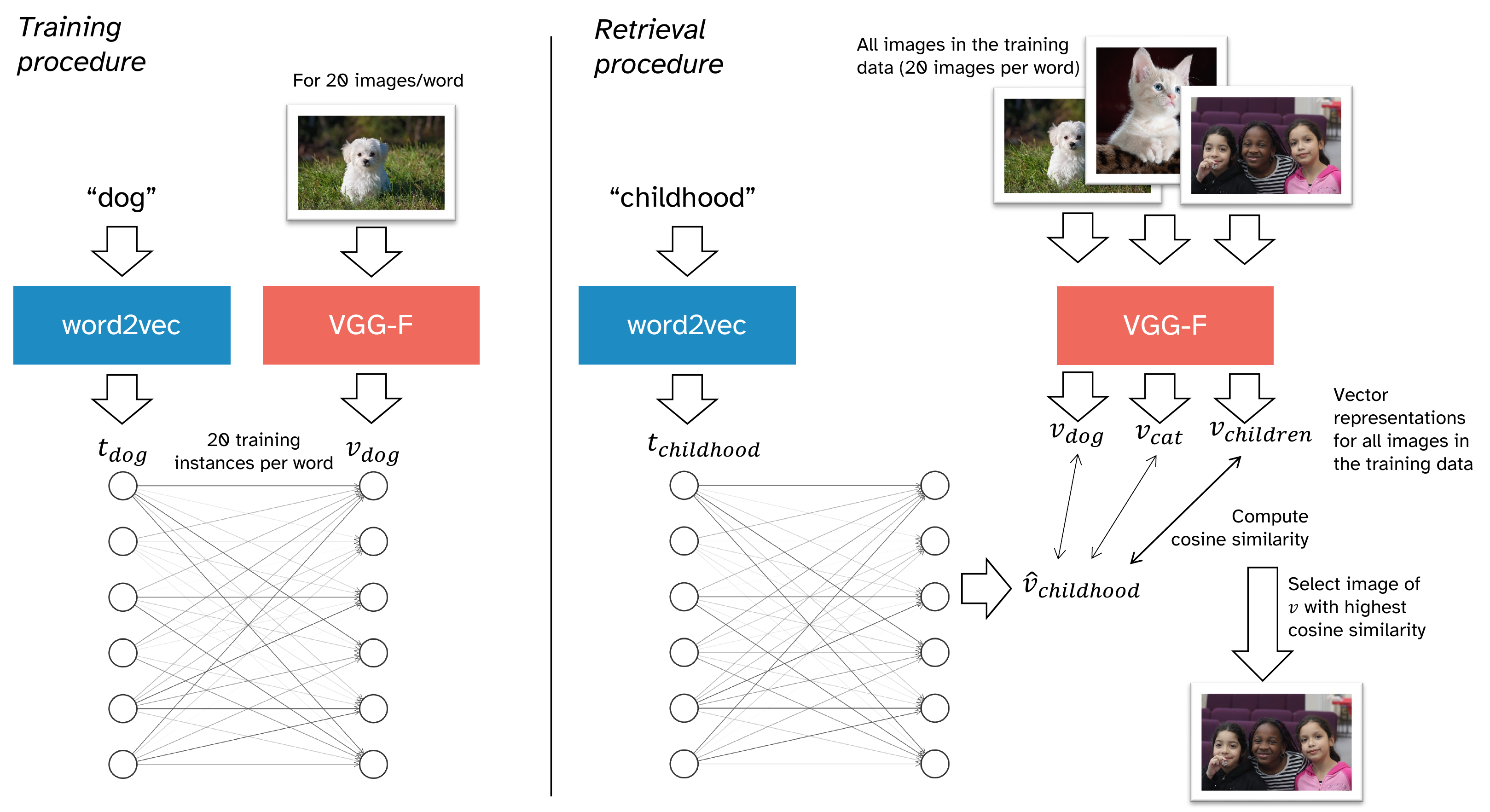}
         \caption{Exemplar model}
         \label{fig:model_exem}
     \end{subfigure}
        \caption{Training and retrieval procedures in the prototype and exemplar models in \citet{Gunther2020Images}. 
        $t$ indicates textual embeddings, $v$ image representations as generated by VGG-F \citep{Chatfield2014ReturnNets} and $p$ prototype image representations. { Images are for illustration purposes only.}}
        \label{fig:model}
\end{figure}

\begin{table}[ht!]
    \centering
    \begin{tabular}{cccc}
    \hline
    condition & target word & predicted image & random control image \\
    \hline
        concrete/maximum & stallion & \includegraphics[height=3cm]{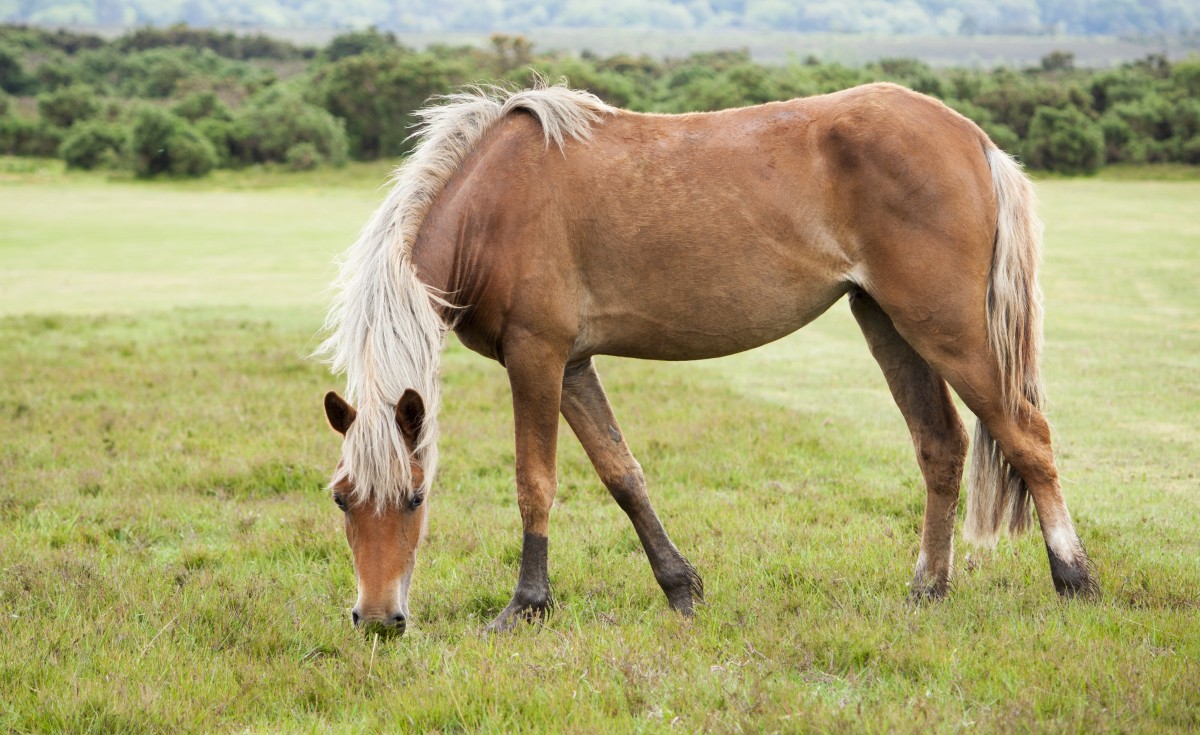} & \includegraphics[height=3cm]{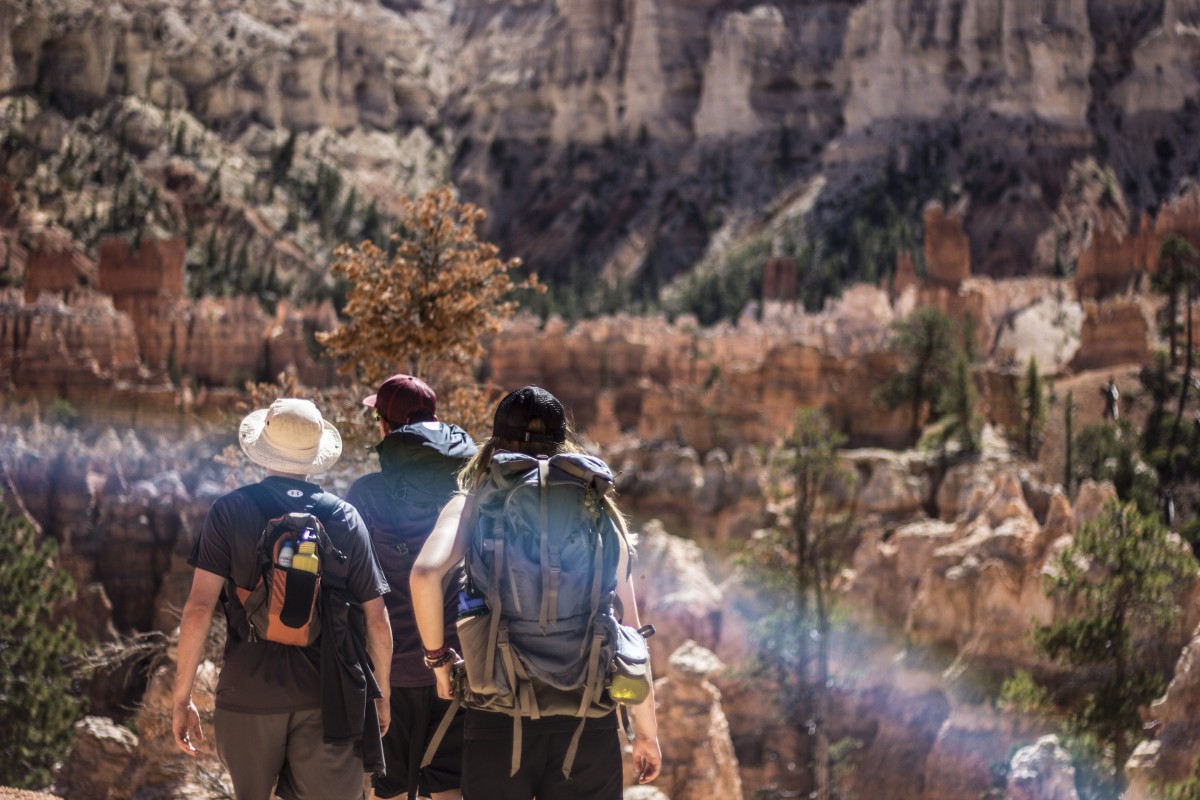} \\
        concrete/near & scout & \includegraphics[height=3cm]{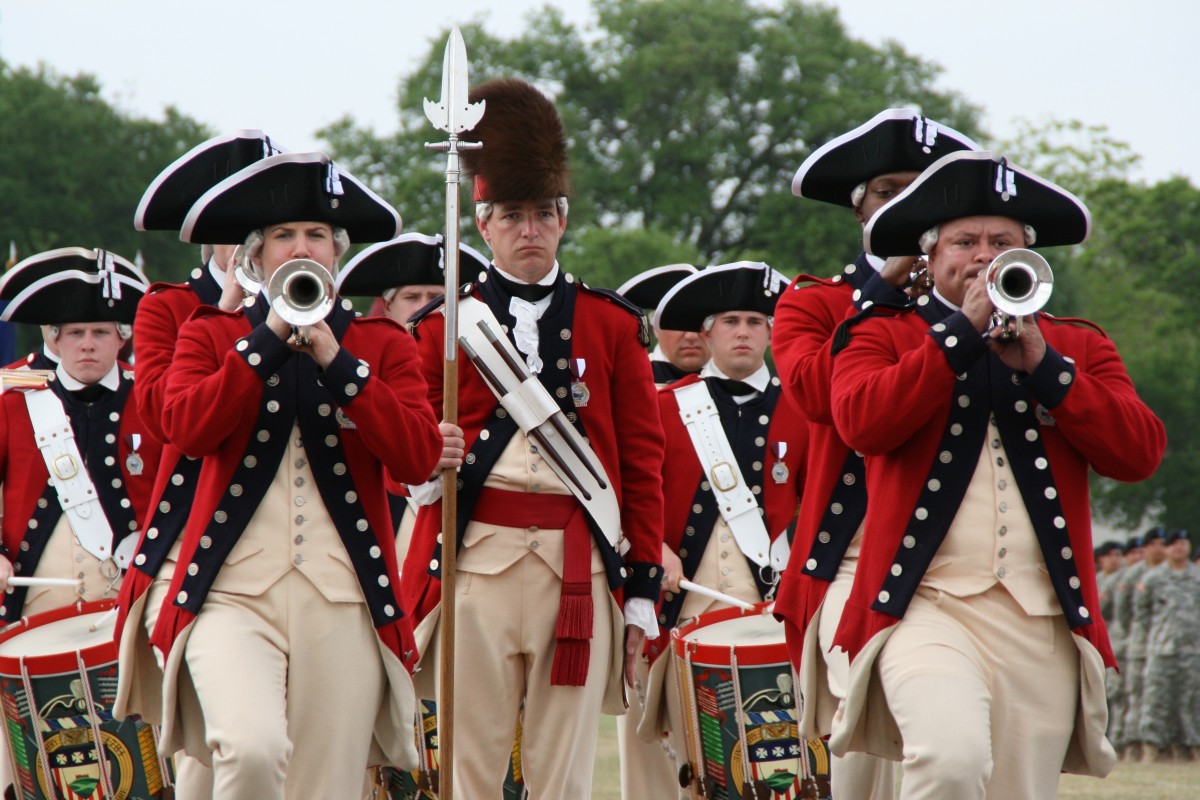} & \includegraphics[height=3cm]{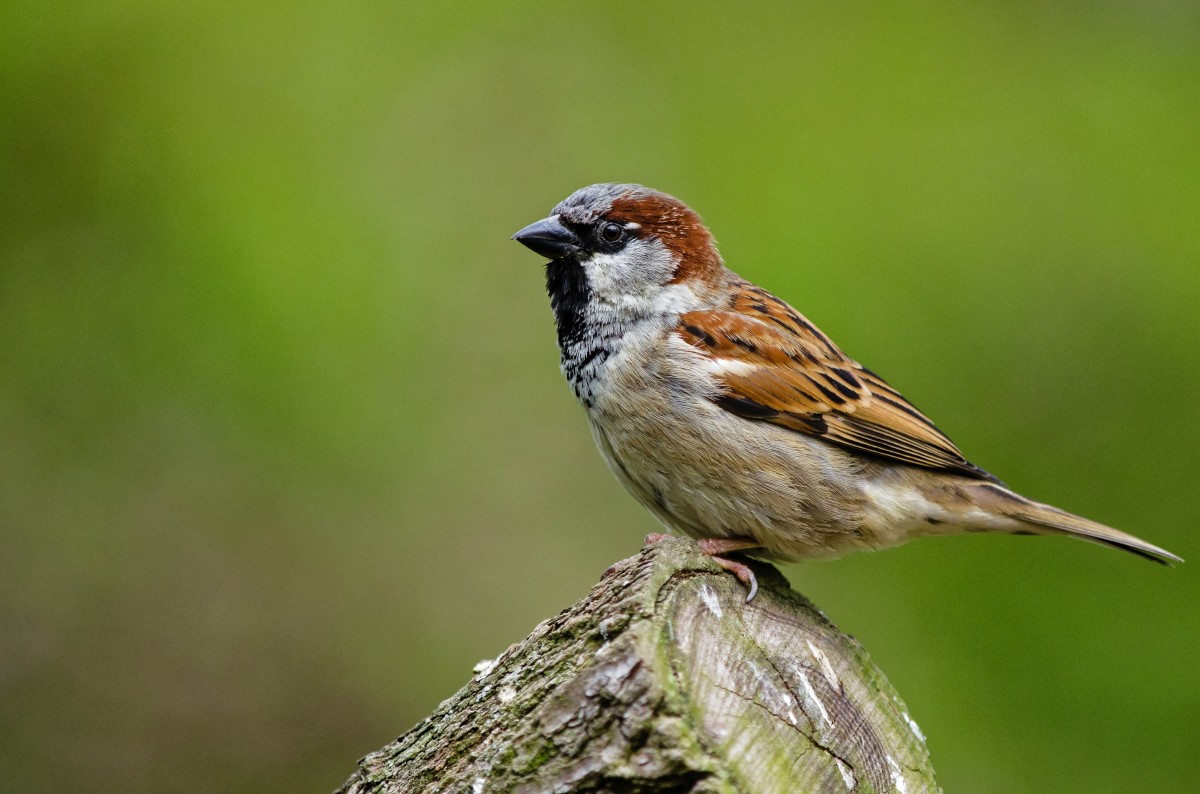}\\
        concrete/far & aspirin & \includegraphics[height=3cm]{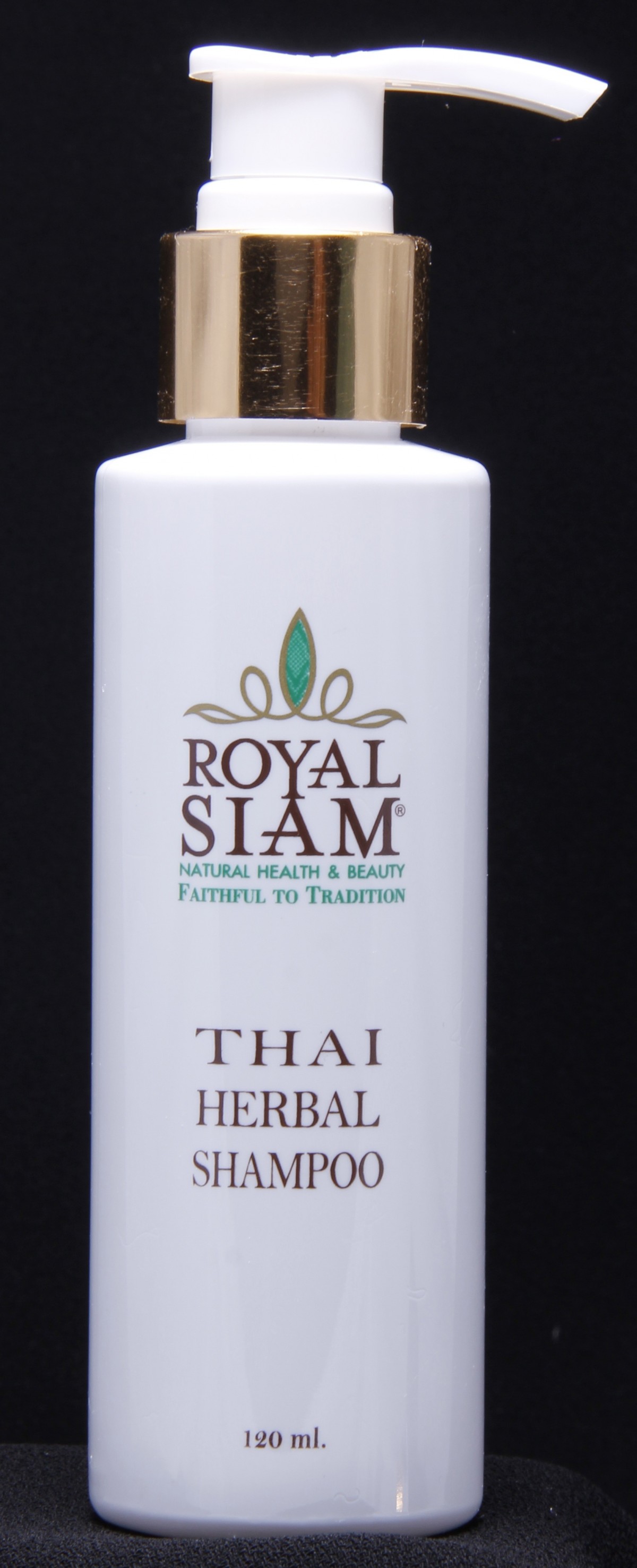} & \includegraphics[height=3cm]{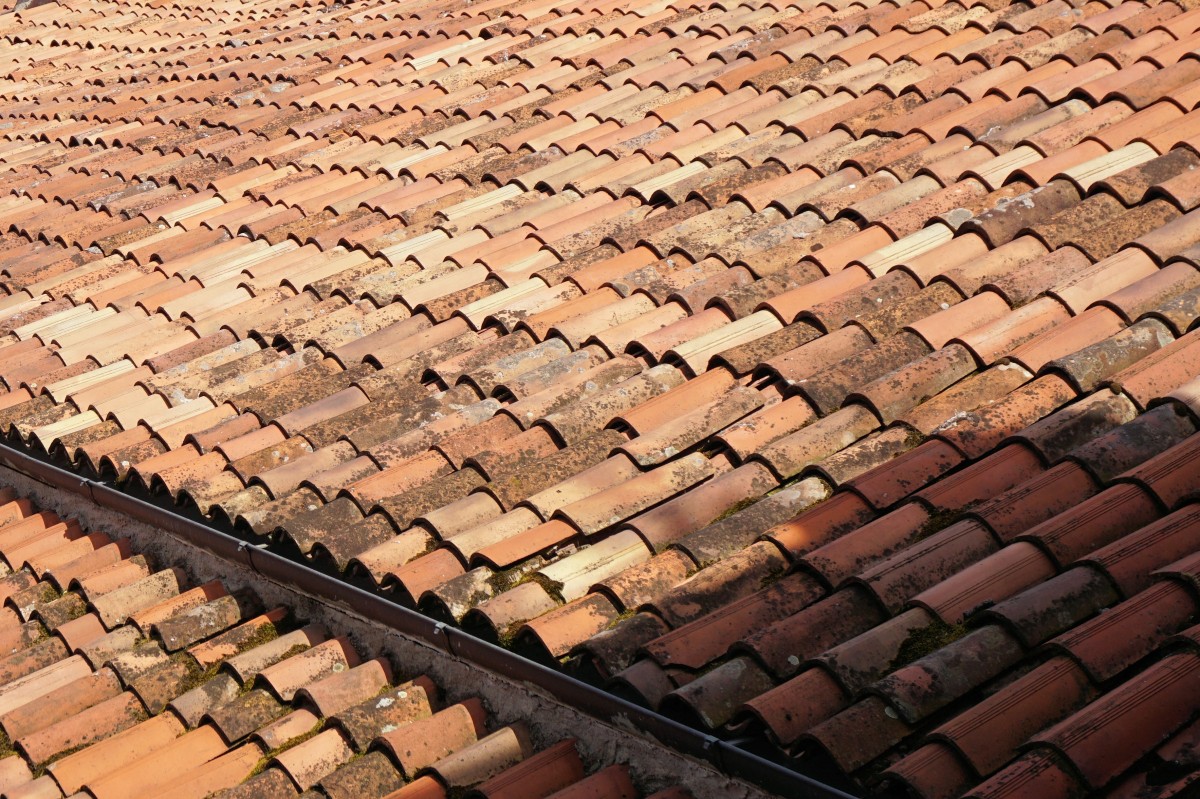}\\
        abstract/near & childhood &\includegraphics[height=3cm]{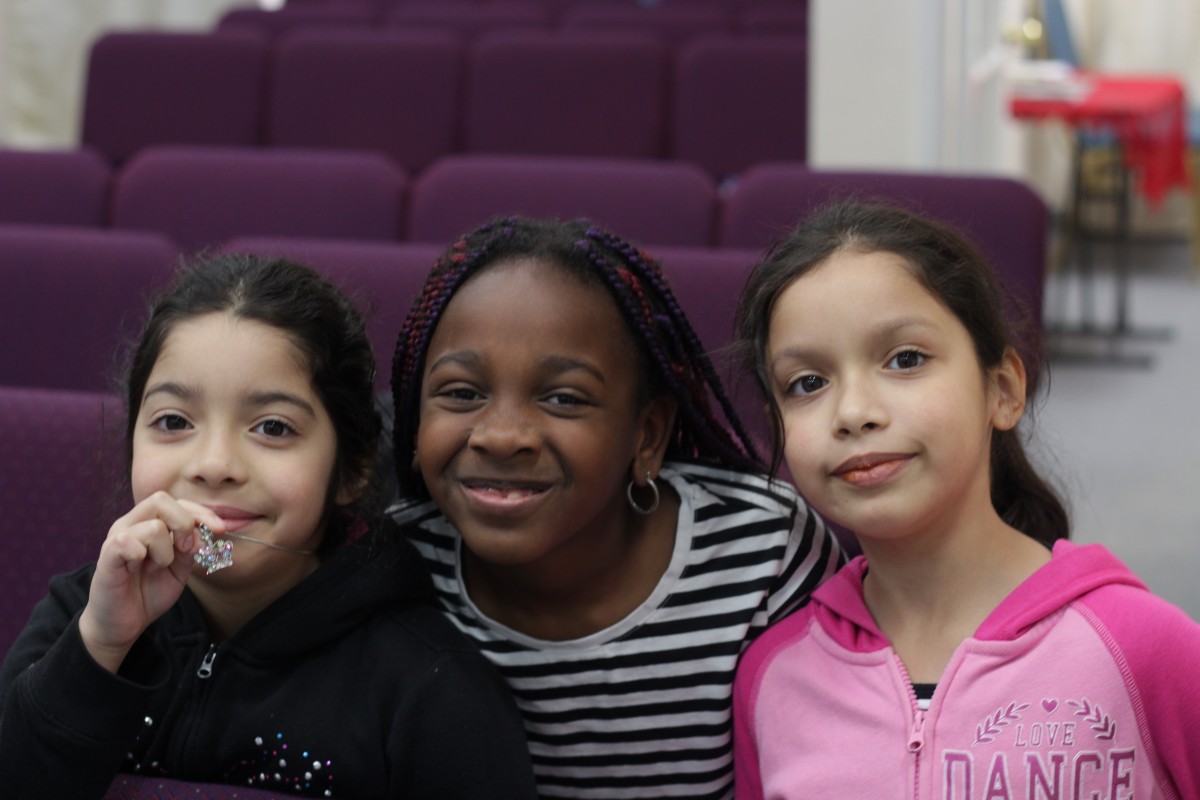} & \includegraphics[height=3cm]{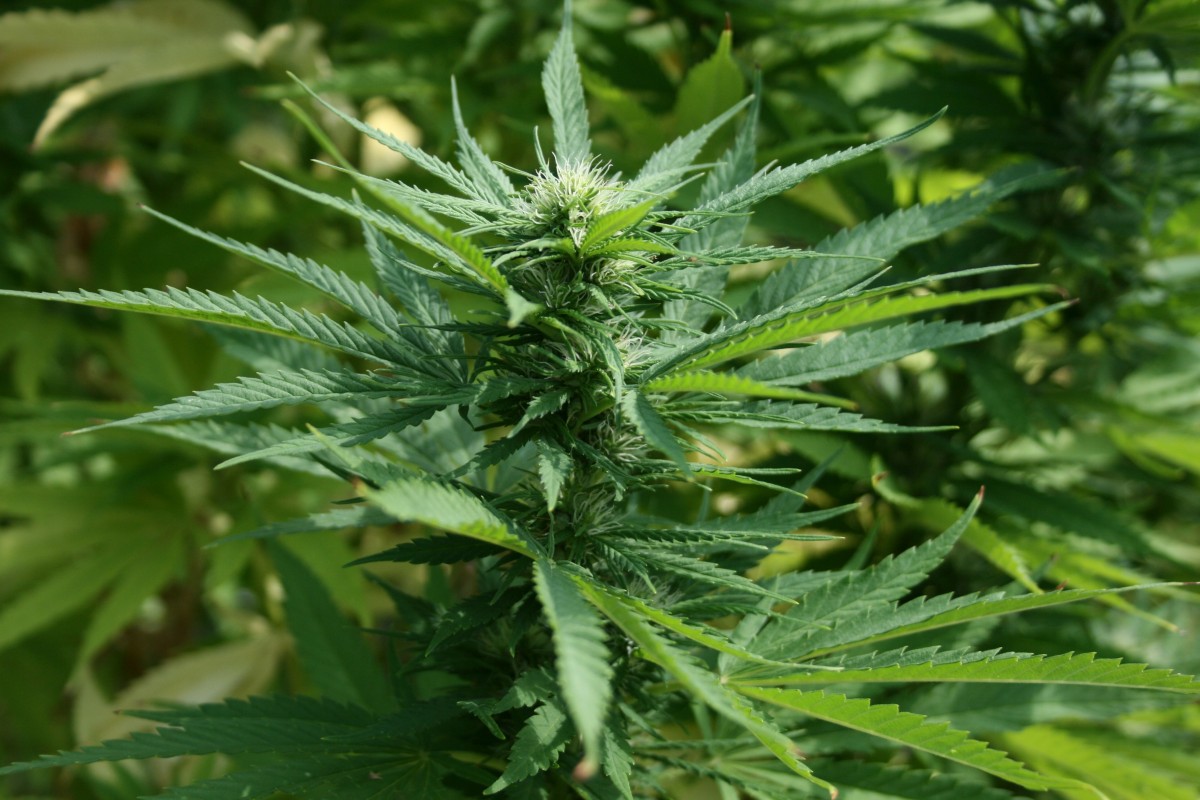}\\
        abstract/far & jealousy & \includegraphics[height=3cm]{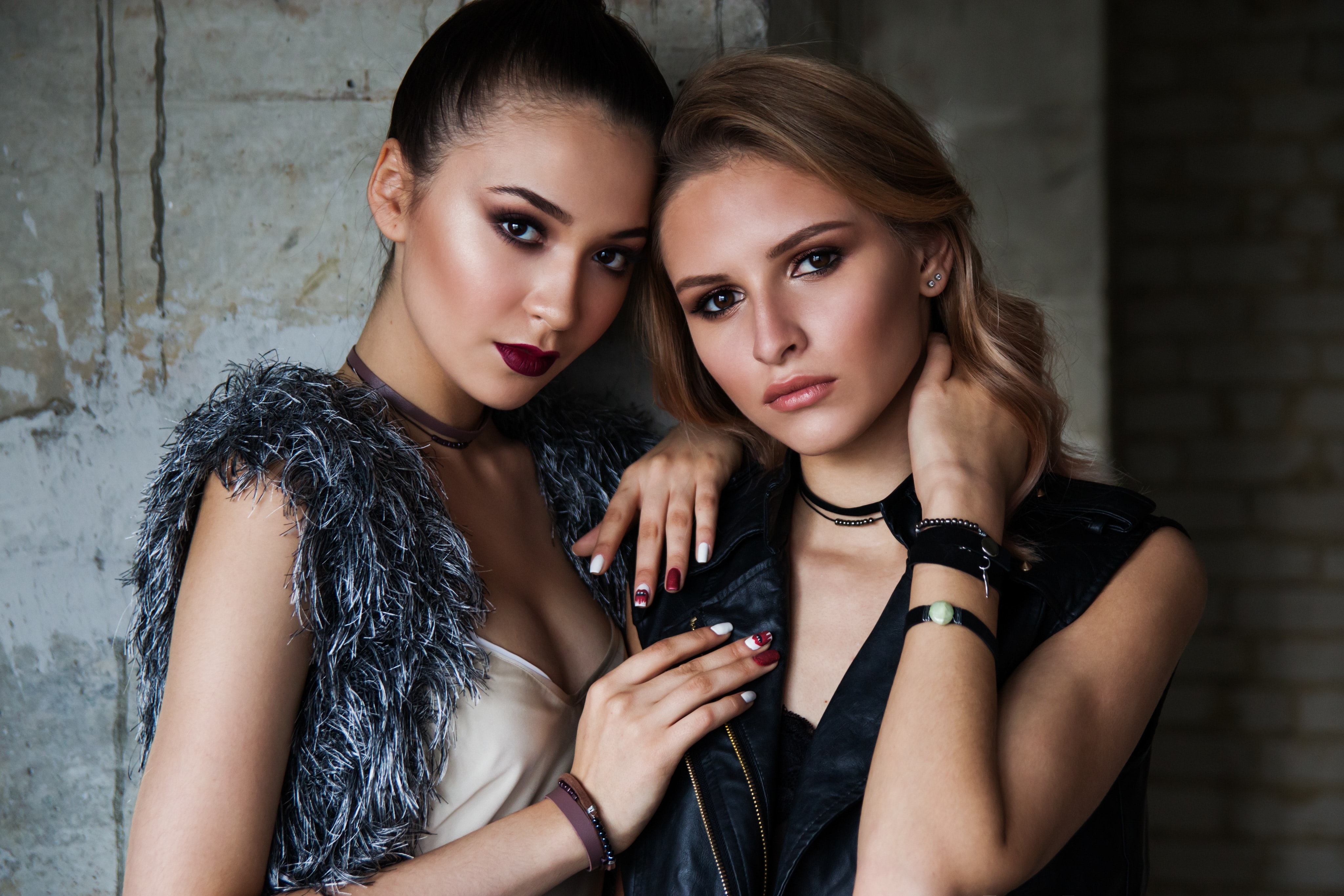} & \includegraphics[height=3cm]{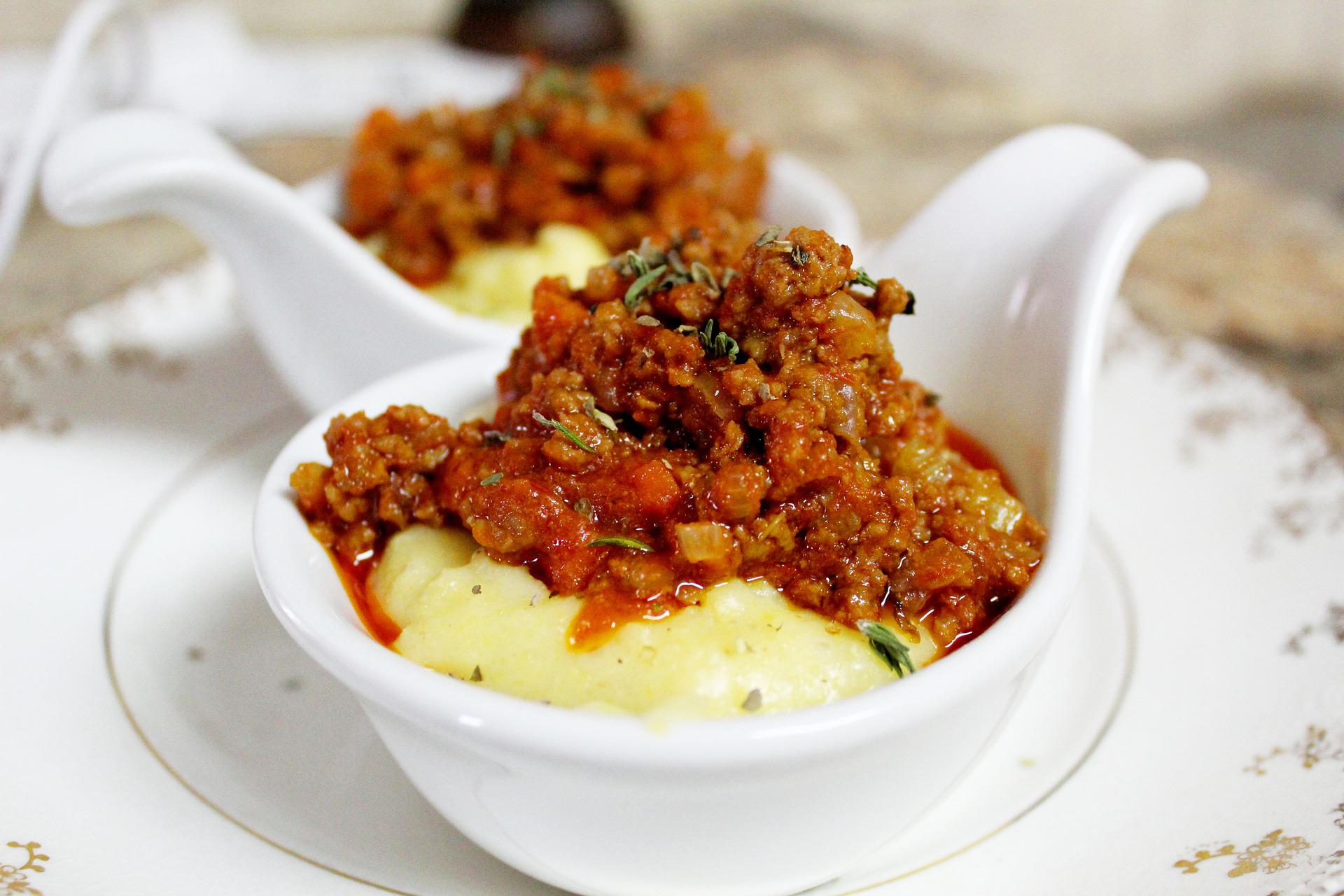}\\
    \hline
    \end{tabular}
    \caption{Examples of predicted (by the prototype model) and random control images for target words from various conditions of \textit{concreteness} and \textit{visual neighbours}. Table adapted from \citet{Gunther2020Images}, all images replaced by visually similar public domain images.}
    \label{tab:gunther_images}
\end{table}


GPVM tested their model by predicting image representations for a range of both abstract and concrete nouns. Since the model is not able to actually generate images, they simply selected existing images whose representations were as close as possible to the predicted image representation. For the exemplar model, they straightforwardly selected the image from the set of training images closest to the predicted image representation. For the prototype model, on the other hand, they first selected the prototype image vector closest to the predicted representation and then selected the training image closest to this prototype image vector. The two training and retrieval strategies are visualised in Figure~\ref{fig:model}. Examples of generated (selected) images can be seen in the center column of Table~\ref{tab:gunther_images}.

In their first two experiments, GPVM 
asked participants to select one of two images for each of the abstract and concrete nouns: either the image predicted by the model, or a random control image (see Table~\ref{tab:gunther_images} for examples of predicted and random control images). Two variables were controlled for: the \textit{concreteness} (\textit{concrete} vs. \textit{abstract}) of the nouns \citep{Brysbaert2014}, as well as the number of semantic neighbours which had an associated image in ImageNet (\textit{visual neighbours}).  \textit{Visual neighbours} are divided into no (\textit{far} condition), few (\textit{near}) and many visual neighbours (\textit{maximum}). 

The participants selected the predicted images above chance level for concrete nouns and abstract nouns with many visual neighbours, and for the exemplar model also for abstract nouns with no visual neighbours. Both \textit{concreteness} and \textit{visual neighbours} are correlated, and they were both found to be predictive of the participant's performance only in the prototype model: the more concrete the noun, and the more \textit{visual neighbours} it had, the more likely the participants were to pick the image predicted by the model. These two variables failed to be predictive for the exemplar model. 
GPVM propose that a possible reason for this is that the exemplar model, being presented with many images for each word instead of just a single average representation as in the prototype model, is able to pick up more ``idiosyncratic visual information'' than the prototype model in the abstract and far conditions, thereby removing any effect of concreteness and visual neighbours. By further disentangling the variables of \textit{visual neighbours} and \textit{concreteness} in a third experiment with the prototype model where both were represented as continuous variables, the authors found that only \textit{concreteness} was significant at predicting participants' accuracy, but not the number of \textit{visual neighbours}. They, therefore, concluded from their experiment that a) it is possible to predict image representations for unseen words from a purely language-based representation even for abstract words, and b) that \textit{concreteness} is a significant, graded predictor of model performance. This result is interesting insofar as the model relies on visual neighbours for predicting images for unseen words but its performance is nevertheless not significantly influenced by the number of visual neighbours.

In the present study, we restricted ourselves to data from experiments 1 and 2. They included 53 and 57 participants, respectively. In both experiments, participants were presented with 115 items (5 conditions with 23 items each). Additionally, they included ten catch trials where the target and random control images were selected manually. Target images were generated by their prototype and exemplar models as described above, while the control images were picked randomly from a set of images not included in the target images. Further details on the experimental materials can be found in \citet{Gunther2020Images}. 

\subsection{Model from \citet{shahmohammadi2022language}}\label{sec:shahmohammadi}
Contrary to  { GPVM}, 
the model from \citeauthor{shahmohammadi2022language} takes the textual context of words into account in the grounding process. As illustrated in Figure~\ref{fig:our_model}, they use image-caption pairs from an image captioning dataset \citep{lin2014microsoft}. This dataset contains images paired with multiple human-annotated descriptions in the form of sentences. For example, instead of having the word \textit{dog} associated with different pictures of dogs, they have access to multiple descriptions of scenes depicting dogs in various situations. Example sentences from their training data are \textit{a dog leaping into the air to catch a frisbee} and \textit{ two dogs poking their heads through curtains at windows}.
In the process of visually grounding, \citeauthor{shahmohammadi2022language} use all the words in such sentences. As a consequence, words are still `aware' of their textual co-occurrence patterns, allowing the model to predict the image vectors with higher fidelity. In what follows, we describe their proposed model in more detail.

For any textual word vector $t_i$ \citep[e.g., from GloVe,][]{Pennington2014Glove:Representation}, the goal is to train a linear alignment/mapping $\textbf{M}$ to visually ground $t_i$ as $g_i = t_i\cdot \textbf{M}$, where $g_i$ denotes the visually grounded version of $t_i$. The alignment $\textbf{M}$ ideally should: a) preserve the essence of the statistical semantics captured from textual corpora, and b) align the textual word vectors with their corresponding visual features in images \citep{shahmohammadi2021learning, shahmohammadi2022language}. Given the properties of linear transformations, the grounded word vector $g_i$ still respects the textual vector space while being informed about its corresponding perceptual properties in images. The Microsoft\_COCO\_2017 dataset \citep{lin2014microsoft} was used to train such an alignment. This dataset is split into 118k train and 5k validation samples, with each sample consisting of a single image along with five different captions describing the image. Unlike many previous approaches \citep{collell2017imagined,Gunther2020Images}, according to which a single textual word vector is mapped into its associated image features, \citeauthor{shahmohammadi2022language} argue that textual context should be taken into account and not discarded, as it contains valuable information about the specific properties of images. For instance, instead of learning a model to map from the word \textit{dog} to an image vector of a dog, their model maps a whole sentence such as \textit{a dog is playing in the grass} into its associated image features. By taking the context \textit{the grass} in the image caption into account, the model is given the opportunity to learn about the grass and to dissociate the dog from the grass. Each word vector is, therefore, aware (connected) of (to) its textual context while being aligned with its visual features. Here, we note that also the images are complex and seldom contain only a single isolated object. { We also note that when presented with photographs, human viewers typically fixate at many different locations \citep{castelhano2008eye}. For complex pictures presented for 12 ms, the mean number of fixations can be as large as 32 \citep{cronin2020eye}.  Instead of taking human image interpretation as wholistic and undifferentiated, we take image understanding to be a rich and diversified process in which multiple objects in an image are scanned and interpreted.
}

For training the linear alignment/mapping $\textbf{M}$, first, all the textual word vectors are piped through the alignment $\textbf{M}$. Then, a contextualized Encoder \citep[e.g., LSTM,][]{hochreiter1997long} produces a single representation vector for the whole sentence while taking all the words and their relative order into account. The contextualized encoder is trained to map the representation of the whole sentence to the image features. In this setup, the encoder provides feedback on how words co-occur and how they should be rearranged to be aligned with the image features. After training the model, the trained alignment $\textbf{M}$ is used to ground all the textual word vectors, including the word vectors of abstract words. That is, given any textual word vector $t_i$, its visually grounded version $g_i$ is easily computed as $g_i = t_i\cdot \textbf{M}$. In this way, they created a visually grounded version of a given textual word embedding model.

\citeauthor{shahmohammadi2022language} evaluated their grounded embeddings on word similarity/relatedness benchmarks, which have been commonly used for evaluation of multi-modal embeddings \citep{park-myaeng-2017-computational, kiros-etal-2018-illustrative,kiela-etal-2018-learning,collell2017imagined}. These benchmarks measure the relatedness/similarity of word pairs based on human-annotated scores.
They used the following datasets for evaluation: WordSim353 \citep{finkelstein2001placing}, MEN \citep{bruni2014multimodal}, SimLex999 \citep{hill2015simlex}, Rare-Words \citep{luong-etal-2013-better}, SimVerb3500 \citep{gerz-etal-2016-simverb}, and MTurk771 \citep{halawi2012large}. Even though each of these  benchmarks evaluates words' meaning from a particular perspective, their grounded embeddings boosted the mean score of GloVe \citep{Pennington2014Glove:Representation} textual embeddings across all benchmarks by almost \textbf{7} percentage points (56.7\% to 63.6\%). They further explored the benefit of grounding in further detail, leveraging from the SimLex999 dataset. This dataset categorizes words into different categories, such as `highly abstract' and `highly concrete'. Their grounding approach is beneficial for both concrete and abstract words and boosts results in other categories (e.g., \textit{verbs} and \textit{adjective}) as well \citep[see][for further details]{shahmohammadi2022language}.

While both \citeauthor{shahmohammadi2022language} and GPVM 
apply a linear mapping to the textual embeddings, the former argues that bridging the gap between language and vision solely using a linear transformation is not ideal. They carried out multiple experiments with increasing complexity in terms of technique and network architecture. They concluded that the right balance is necessary to obtain high-quality embeddings that perform well on word similarity tasks. Their simplest approach, which they refer to as `Word-Level', is similar to the model by GPVM 
in which textual word vectors of words in isolation (e.g., \textit{dog}) are mapped to corresponding image vectors through a linear transformation. The grounded embeddings are then constructed by mapping the textual word vectors through the trained linear mapping. Evaluation on various word similarity benchmarks  \citep[e.g., MEN,][]{bruni2014multimodal}, revealed that embeddings obtained in this way underperform severely (by $> 30$ percentage points) compared to purely textual embeddings. {Applying very deep and complex models (e.g., applying multiple layers of LSTMs or transformer-based models \citep{kenton2019bert} for the mapping $\textbf{M}$) also resulted in a drop in performance.} The right balance in terms of the depth and complexity of the network is therefore necessary. For further implementation details about their model and analysis, see  \citet{shahmohammadi2022language}.

\subsection{Procedure}
\label{sec:procedure}

\begin{figure}
    \centering
    \includegraphics[width=\textwidth]{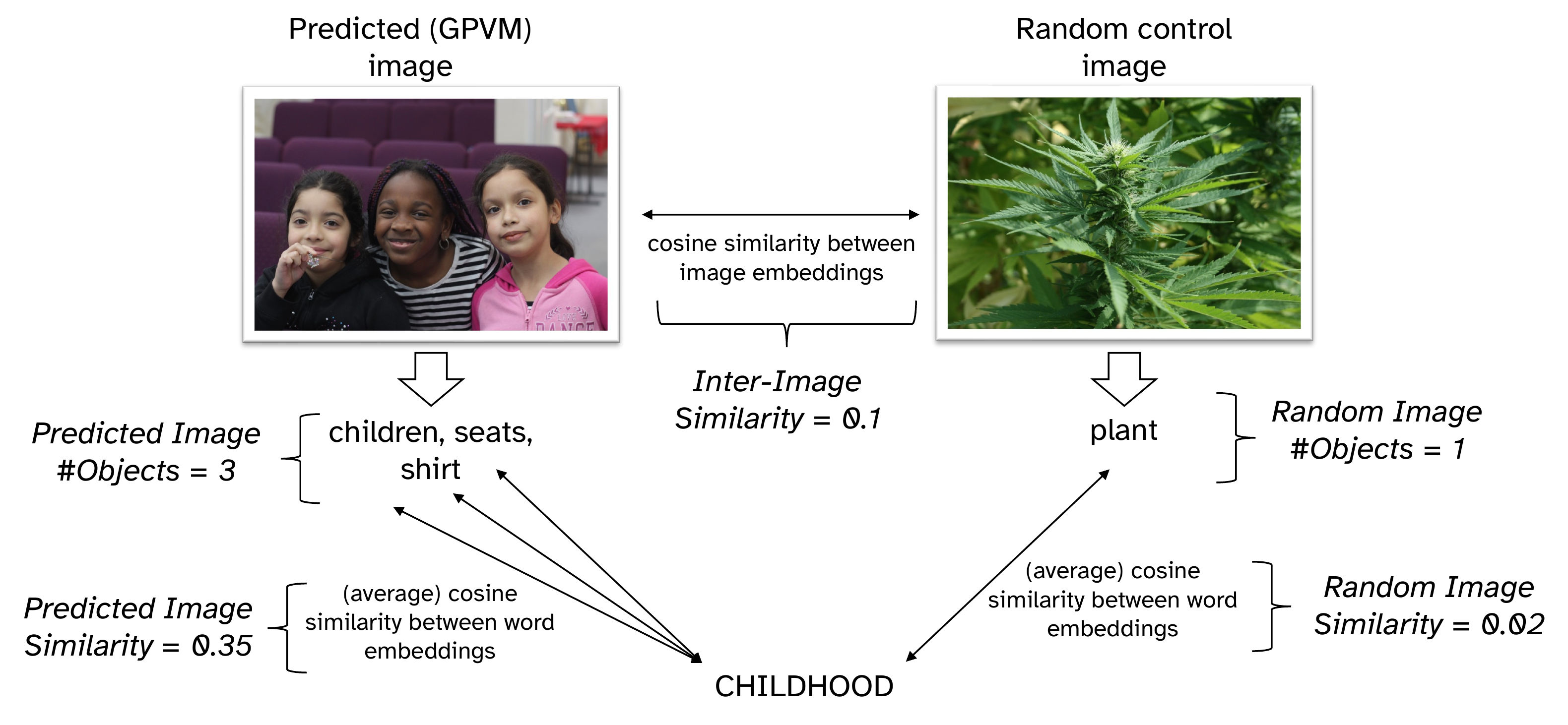}
    \caption{In \citet{Gunther2020Images}, participants are presented with two images (predicted and random control image) and a target word (here: childhood). They then have to select the image which better fits the target word. We calculate five measures: First, we use a CNN to automatically extract the object names visible in the predicted and random images and count them ({\textit{Predicted Image \#Objects} and \textit{Random Image \#Objects}}). Then we compute the average cosine similarity between the embeddings of the target word and the object names respectively ({\textit{Predicted Image Similarity} and \textit{Random Image Similarity}}). We also calculate the cosine similarity between the image embeddings of the two images {\textit{(Inter-image similarity)}}.}
    \label{fig:task_analysis}
\end{figure}

In order to clarify whether participants associated the target noun with object names depicted in the images rather than generating mental images and basing their decisions on comparisons of these images, we first extracted the names of the objects (or labels) in both the predicted image and the random control image, using a pre-trained CNN model \citep{tan2019efficientnet}. 
{
We then used the object names to obtain the corresponding word embeddings.  In other words, our goal here is to retrieve the semantics of the objects in the images, and \textit{not} to extract their orthographic written forms.  The cognitive process that we are approximating with an engineering solution is the process of understanding what the objects in an image are.  Note that empirical studies using eye-tracking to trace image interpretation show that images are typically scanned with many fixations at many different image locations \citep{castelhano2008eye,cronin2020eye}.  
}

More specifically, for each image, we extracted the names of the top 10 classes\footnote{The CNN classes include the true classes for all the images used by GPVM 
as they both utilized the same image database.} predicted by the CNN model. Examples of predicted classes for a particular image are \textit{`bagel', `plate', `pretzel', `dough', `bakery', `butternut\_squash', `cheeseburger', `spaghetti\_squash', `chocolate\_sauce'}, and \textit{`acorn\_squash'}. Here, the underscore represents a space character. For most of these very specific subcategories, no embeddings are available. As a result, the number of objects detected by our algorithm closely follows the number of objects in the images.

Then we modelled participants' behaviour using two approaches. 
In our first, very simple baseline approach (called ``Max'' in the remainder of this paper), we used the average cosine similarity 
{
between the embedding of the query (i.e., target) word and the embeddings of all the object embeddings detected in the predicted image, and in the control image, presented to a participant at a given trial. The average cosine similarity for a given image provides a measure of the likelihood of selecting that image for the given query word. This resulted in two measures:}
\begin{itemize}
    \item \textit{Predicted Image Similarity}: The mean cosine similarity of the target word's embedding with the embeddings of the objects in the image predicted by {the model of { GPVM} (henceforth the GPVM image);}
    \item \textit{Random Image Similarity}: The mean cosine similarity of the target word's embedding with the embeddings of the objects detected in the random control image.
\end{itemize}
To model human selection behavior, the image with the higher image similarity was selected as our model's choice. For example, if in trial 1, \textit{Predicted Image Similarity} was higher than \textit{Random Image Similarity}, we selected the GPVM image. We did this for the prototype and exemplar models separately. 
{
This cognitively rich model of how participants solve the experimental task contrasts with the lean, vision-only model of GPVM, who assume that a target query word makes contact with its corresponding embedding. That, in turn, generates an internal image (using a pre-trained mapping from word embeddings to images) that is subsequently compared with the two images presented to a participant, without any further involvement of higher cognitive processes evaluating what the objects present in images actually are.
}

In our second approach (in the following called GAM), we used the following additional predictors: 
\begin{itemize}
    \item \textit{Inter-Image Similarity}: The mean cosine similarity between the GPVM image and the random control image vectors;
    \item \textit{Predicted Image \#Objects}: The number of object labels (e.g., dog, tree, \ldots) in the GPVM image for which word embeddings were available in the set of embeddings;
    \item \textit{Random Image \#Objects}: The number of objects labels in the random control image for which word embeddings were available in the set of embeddings.
\end{itemize}
An overview of the extracted measures can be found in Figure~\ref{fig:task_analysis}. Additionally, we used the two predictors provided by GPVM, 
which capture the number of \textit{visual neighbours} (Distance) and  \textit{concreteness} (WordType). All of these metrics were used as predictors in a Generalised Additive Model (GAM) with a logistic link function, as implemented in the \textit{mgcv} package \citep{wood2011gam} in R. GAMs can model non-linear relationships between independent and dependent variables. We tested a range of different GAMs with these predictors, which we compared using the Akaike Information Criterion (AIC). While models with interactions between predictors (using tensor product smooths) gave a substantially better fit to the data, they also proved to be much harder to interpret. Therefore, we selected GAMs with main effects only for both the prototype and the exemplar models. We compared two sets of GAMs inspecting both AIC and predictions, one based on grounded vectors and one based on purely textual embeddings.  \footnote{Generated measures and analysis notebooks can be found in the Supplementary Materials at \url{https://osf.io/7rxde/}.}

\section{Results}\label{sec:results}

\subsection{Q1: Can we model participant behaviour without assuming participants generate mental images?}\label{sec:q1}

\subsubsection{Max models}

{
Our hypothesis is that participants 
compare the meanings of the objects in the two images with the meaning of the target noun (using the respective embeddings),} and select the image with the higher similarity. This idea is operationalized in our Max approach. We test Max using both textual and grounded GloVe and a version of (textual and grounded) Word2Vec \citep{mikolov2013efficient} that was used by GPVM. 
The first minimal evaluation criterion for our approach is that it is able to differentiate the GPVM image from the random control image with a higher-than-chance probability. We found that this is indeed the case for the Max approach based on both textual and grounded, GloVe and Word2Vec embeddings, and prototype and exemplar setups (proportions test; $p<0.0001$). This shows that our approach provides at least a theoretical possibility of how to solve the task.

Secondly, to investigate how well the Max approach approximates human behaviour, we measure the proportion where Max selected the GPVM image. We can thus view our four embedding types as \textit{virtual participants}. If our hypothesis for predicting participants' selection behaviour holds true, we expect the virtual participants to show a similar preference (compared to participants' preference) for the GPVM image. 

The results of our experiments are reported in Table~\ref{tab:virtual} for both the exemplar and prototype setups. We observe that the mean scores of the virtual participants are quite close to the mean scores of real participants (i.e., ``Participants'' in Table~\ref{tab:virtual}). The absolute difference between the mean score of the virtual participants and that of the participants is reported and labeled as $\Delta$ in the Table. Lower $\Delta$ values indicate a better fit for modeling the participants’ preferences.

\begin{table}[ht]
\begin{subtable}[h]{\textwidth}
\centering
\begin{tabular}{lcccccc}
\hline \textbf{Embeddings} & \textbf{A. far} & \textbf{A. Near} & \textbf{C. Far} & \textbf{C. Near} & \textbf{C. Max} & \textbf{Mean ($\Delta$)} \\ \hline

Max: GloVe          & 82.61 & 69.57 & 56.52 & 90.91 & \bftab 86.96 & 77.31 (07.06) \\
Max: ZSG-GloVe  & \bftab 52.17 & \bftab 60.87 & \bftab 69.57 &  81.82 & 91.30 & \bftab 71.15 (00.90)  \\
Max: W2V            & 65.22 & 73.91 & 78.26 & \bftab 86.36 & \bftab 86.96 &  78.14 (07.89)  \\
Max: ZSG-W2V & 65.22 &  78.26 & 73.91 &  90.91 & 91.30 & 79.92 (09.67)  \\
\hline
Participants  & 52.00 & 64.00 & 66.00 & 84.25 & 85.00 & 70.25   \\
\hline
\end{tabular}
\caption{Prototype model}
\label{tab:virtual_proto}
\end{subtable}

\begin{subtable}[h]{\textwidth}
\centering
\begin{tabular}{lcccccc}
\hline \textbf{Embeddings} & \textbf{A. far} & \textbf{A. Near} & \textbf{C. Far} & \textbf{C. Near} & \textbf{C. Max} & \textbf{Mean  ($\Delta$)} \\ \hline

Max: GloVe          &  65.22  &  91.30 & \bftab 73.91 & \bftab 77.27 &91.30 &79.80 (07.40)  \\
Max: ZSG-GloVe   &69.57 & \bftab 86.96 &65.22 &68.18 &86.96 &  75.38 (02.98) \\
Max: W2V            &73.91 & \bftab 86.96 &60.87 & \bftab77.27 &\bftab 78.26 & 75.45 (03.05)    \\
Max: ZSG-W2V & \bftab 60.87 & \bftab 82.61 & \bftab 60.87 & 72.73 & 86.96 & \bftab 72.81 (00.41)  \\
\hline
Participants  &62.00 &74.00& 73.00 &76.00&77.00&72.40  \\
\hline
\end{tabular}
\caption{Exemplar model}
\label{tab:virtual_exem}
\end{subtable}
\caption{Modeling participants' preference for the predicted images. The numbers in the top section of each table represent the percentage of trials in which each virtual participant chooses the GPVM image over the random image. The numbers in the last row of each table show the mean percentage of trials in which the GPVM image was selected, averaged over all human participants. The best results are marked in bold in each category. The absolute difference between the mean score of our model's prediction and that of the participants is reported and labeled as $\Delta$ in the Table. Lower $\Delta$ values indicate a better fit for modeling the participants' preferences.}
\label{tab:virtual} 
\end{table}

\begin{figure}[ht!]
     \centering
     \begin{subfigure}[b]{.45\textwidth}
         \centering
         \includegraphics[width=\textwidth]{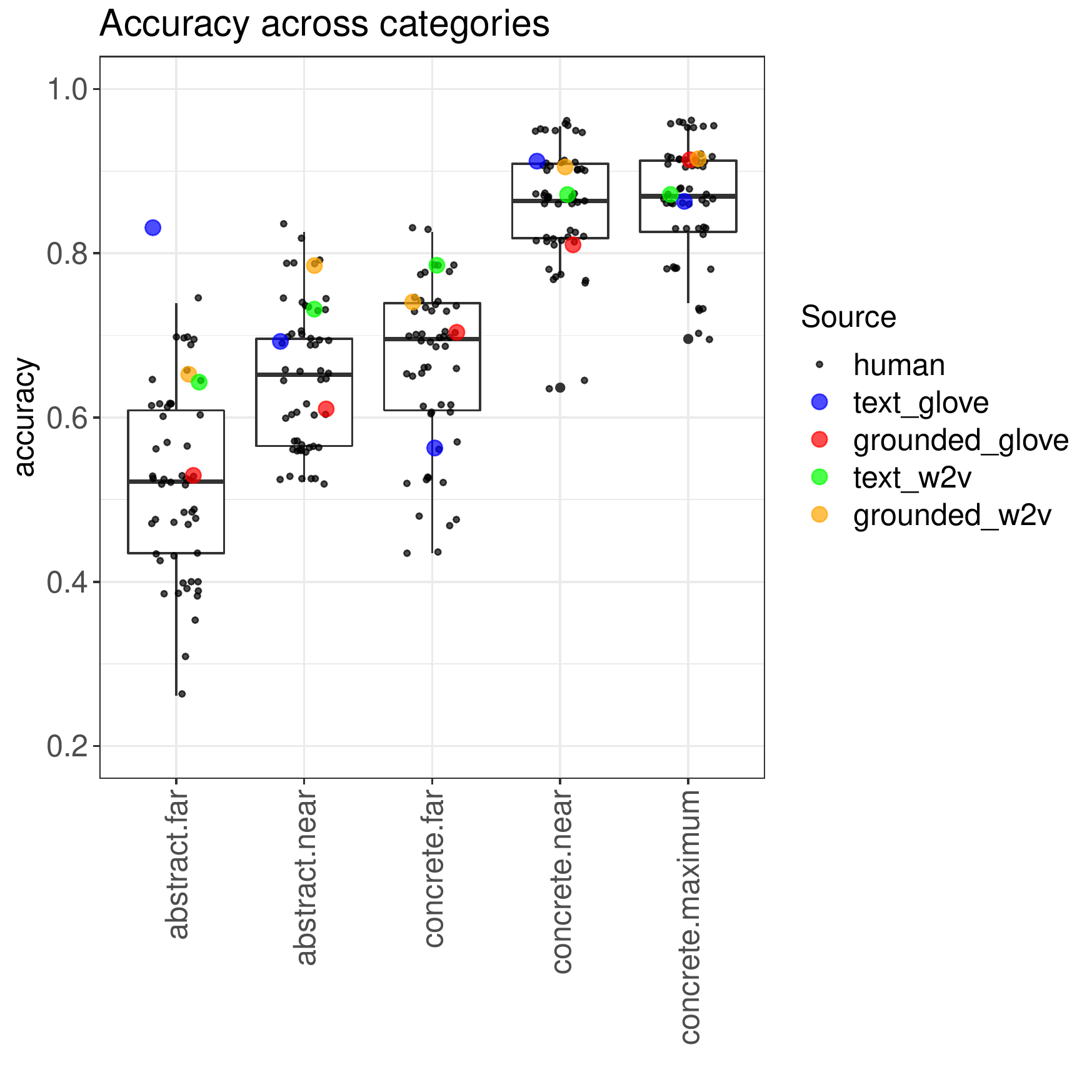}
         \caption{Prototype model}
         \label{fig:acc_proto}
     \end{subfigure}\hfill
     \begin{subfigure}[b]{.45\textwidth}
         \centering
         \includegraphics[width=\textwidth]{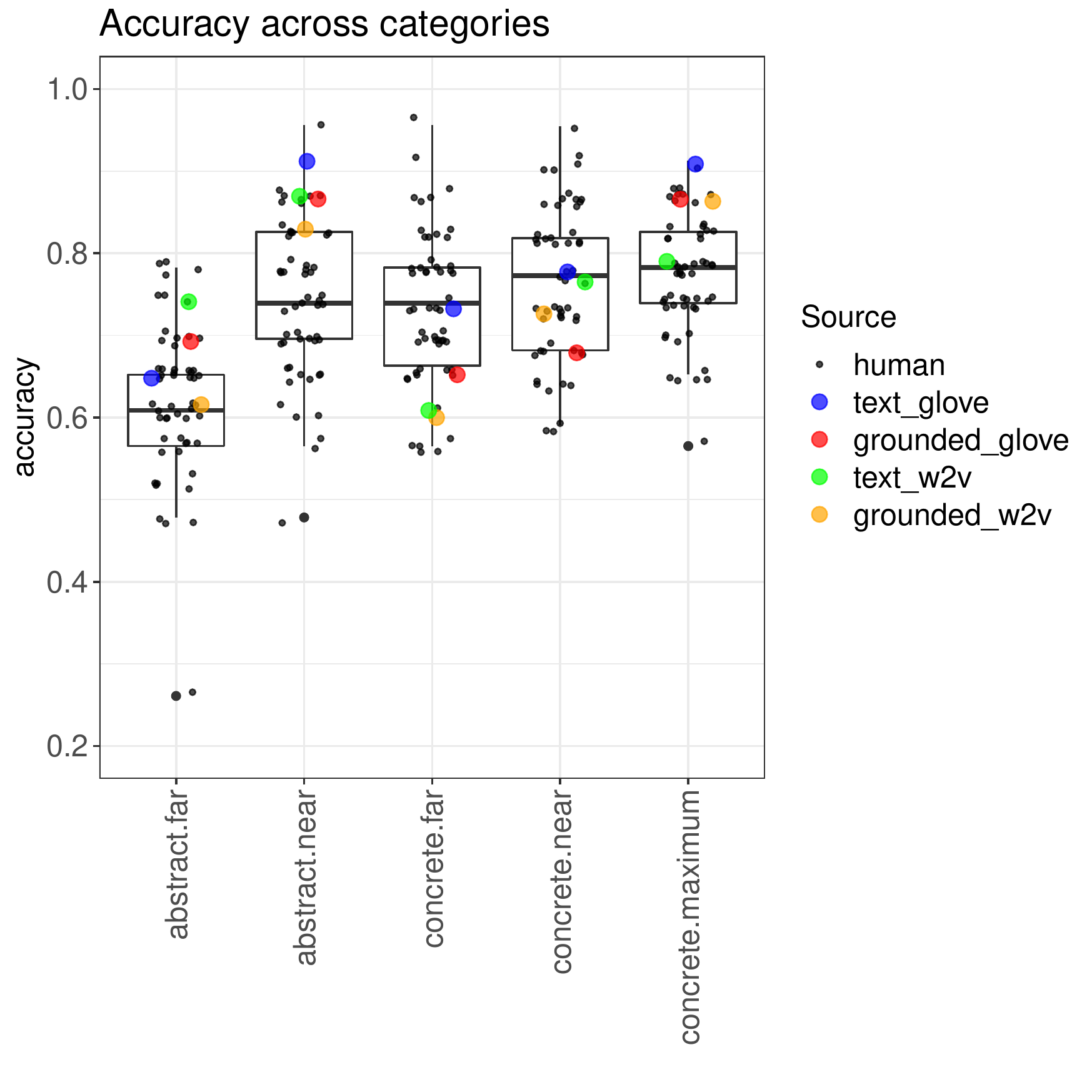}
         \caption{Exemplar model}
         \label{fig:acc_exem}
     \end{subfigure}
        \caption{Performance of the four embedding types compared to human participants in \citeauthor{Gunther2020Images} for the prototype and exemplar setup. Boxplots are based on human data points only. The performance of most embedding types is well within the range of human participants.}
        \label{fig:acc}
\end{figure}

Viewing our four embedding types as ``virtual participants'' begs the question of whether their performance fits into the distribution of human performance in GPVM. 
Figure~\ref{fig:acc} shows that the performance of most of the embedding types falls well within plausible participant performance across all categories of WordType and Distance. The clearest outlier is the model with textual GloVe embeddings in the abstract far category for the prototype setup. We will return to the question of grounded vs. textual embeddings and differences across concreteness/visual neighbour conditions below.

Thirdly, we expect our virtual participants to show the same effects of WordType and Distance as human participants. Unfortunately, the low number of data points per embedding type (114) made running individual, embedding-specific logistic regression models akin to the one by GPVM 
for human participants impossible. However, we do note that the (all non-significant) effects pointed in the same directions as for human participants for all embedding types for the prototype setup: both higher concreteness and more Distance lead to a higher probability of selecting the GPVM image. When combining the data of all four embedding types into one logistic regression model (without by-embedding random effects, since they prevented the model from converging), the effects still pointed in the same directions and were significant. In the exemplar setup,  GPVM 
did not find a significant effect of any of the conditions other than a significant intercept (indicating that human participants generally performed above chance). The combined logistic regression model for all four embedding types based on the exemplar setup showed no effect of concreteness which is in line with findings by GPVM 
but did find a positive effect of Distance which was not found in GPVM
(models in Supplementary Materials).

Fourthly, we can use the predictions of the Max model to directly predict participants' behaviour. The results for the prototype experiment are reported in Table~\ref{tab:res_proto}. The bottom rows of both tables report the proportion of trials  in which participants select the GPVM image over the random control image across categories in the two experiments by GPVM. 
Participants tended to select the GPVM image, the more concrete the target words were and the more visual neighbors they had (see also Figure~\ref{fig:acc}). Since 
here we are interested in predicting
participant behaviour rather than trying to differentiate the GPVM image and random control image, the numbers in the upper and lower parts of the two tables cannot be compared directly. Rather, we want our models to score as closely to 100\% accuracy as possible, indicating a complete matching to participants' selections. Focusing on the mean performance across all concreteness/distance categories, Table~\ref{tab:res_proto} shows that participants' preferences are predicted fairly well (Mean accuracies for Max models are 68\%, 70\%, 69\%, and 71\% for textual and grounded GloVe and textual and grounded Word2Vec). Table~\ref{tab:res_exem} presents the results for the exemplar experiment. Mean accuracy for Max models tends to be somewhat lower compared to the prototype setup (70\%, 66\%, 67\%, and 66\% for textual and grounded GloVe and textual and grounded Word2Vec).

\begin{table}[ht]
\begin{subtable}[h]{\textwidth}
\centering
\begin{tabular}{lcccccc}
\hline \textbf{Embeddings} & \textbf{A. Far} & \textbf{A. Near} & \textbf{C. Far} & \textbf{C. Near} & \textbf{C. Max} & \textbf{Mean} \\ \hline

Max: GloVe          & 52.56 & 62.66 & 56.44 & 84.74 & 83.46 & 67.97 \\
Max: ZSG-GloVe    & \bftab 61.41 & 64.52 & 60.79 & 76.79 & \bftab 87.5 & 70.20 \\
GAM: GloVe & 45.42 & \bftab71.35 & 61.41 & \bftab 87.99 & 85.02 & 70.24\\
GAM: ZSG-GloVe           &  53.96 & 69.49 & 62.03 & \bftab87.99 & \bftab 87.5 & \bftab 72.19\\
Max: W2V            & 54.27 & 66.54 & \bftab 67.16 & 81.33 & 77.72 & 69.40   \\
Max: ZSG-W2V  & 55.82 &  67.93 & 58.93 &  87.34 & \bftab 87.5 & 71.05   \\
GAM: W2V & 49.77 & 66.54 & 60.95 & \bftab 87.99 & 85.79 & 70.21\\
GAM: ZSG-W2V  & 55.98 & 65.45 & 62.19 & \bftab 87.99 & \bftab 87.5 & 71.82\\
\hline
Participants  & 52.00 & 64.00 & 66.00 & 84.25 & 85.00 & 70.25   \\
\hline
\end{tabular}
\caption{Prototype model}
\label{tab:res_proto}
\end{subtable}

\begin{subtable}[h]{\textwidth}
\centering
\begin{tabular}{lcccccc}
\hline \textbf{Embeddings} & \textbf{A. Far} & \textbf{A. Near} & \textbf{C. Far} & \textbf{C. Near} & \textbf{C. Max} & \textbf{Mean} \\ \hline

Max: GloVe          &57.71  &  72.73 & 68.30 &70.33 &81.5 & 70.11 \\
Max: ZSG-GloVe   &49.09 &66.17 &65.45 &73.00 &78.58 &66.45 \\
GAM: GloVe      & 58.02 & 72.73 & 73.28 & 74.63 & 81.50 & \bftab 72.03 \\
GAM: ZSG-GloVe & \bftab 62.85 & 68.93 & \bftab74.31 & \bftab75.87 & 73.7 &  70.94\\
Max: W2V            &55.81 &64.58 &63.00 &74.00 &75.42& 66.56    \\
Max: ZSG-W2V  &52.73 &63.40 &63.48 &73.22& 76.36 &65.83   \\
GAM: W2V & 55.77 & \bftab 69.09 & 69.25 & 75.54 & \bftab 81.11 & 70.95\\
GAM: ZSG-W2V  & 57.23 & 67.27 & 70.91 & 75.21 & 79.13 & 70.00\\
\hline
Participants  &62.00 &74.00& 73.00 &76.00&77.00&72.40  \\
\hline
\end{tabular}
\caption{Exemplar model}
\label{tab:res_exem}
\end{subtable}
\caption{Evaluation of two textual embeddings and their grounded versions on the behavioural experiment by \cite{Gunther2020Images}. 
The numbers in the top section of each table represent the percentage of trials in which the models correctly predict the participants' choices. The numbers in the last row of each table (labelled `Participants') show the mean percentage of trials in which the GPVM image was selected, averaged over all human participants.
A. and C. indicate abstract and concrete words respectively. Far, near and max refer to the distance of visual neighbours.}
\label{tab:marco} 
\end{table}

We can therefore conclude that it is indeed possible to model the behavioural experiment by GPVM 
without assuming that participants generate mental images and that even meaning representations based on textual information only are able to model participants' behaviour fairly well.

\subsubsection{GAM models}

Thus far, our predictions for participants' selection preferences have been based on average similarity scores for the images.   Accuracy can be improved by also taking into account the similarity of the GPVM images and their controls, the number of objects in these images, as well as the two factorial predictors considered by GPVM: 
Distance and WordType.  As mentioned above, we use logistic GAMs to obtain predictions for participants selection decisions. Using GAMs also enables us to investigate what effects \textit{Predicted Image Similarity} and \textit{Random Image Similarity} have on participant behaviour. In the prototype setup, accuracy improved for all pairs of comparisons. For instance, the accuracy of Max: ZSG-GloVe,  70.20, improved for  GAM: ZSG-GloVe to an accuracy of 72.19.

The GAM (see Table~\ref{tab:gam_proto}) indicated that prototype images for concrete words were more often selected compared to abstract words, and that images for words with more image neighbors were also selected more often.  These effects mirror those observed by GPVM. 

\begin{figure}[ht!]
     \centering
     \begin{subfigure}[b]{\textwidth}
         \centering
         \includegraphics[width=\textwidth]{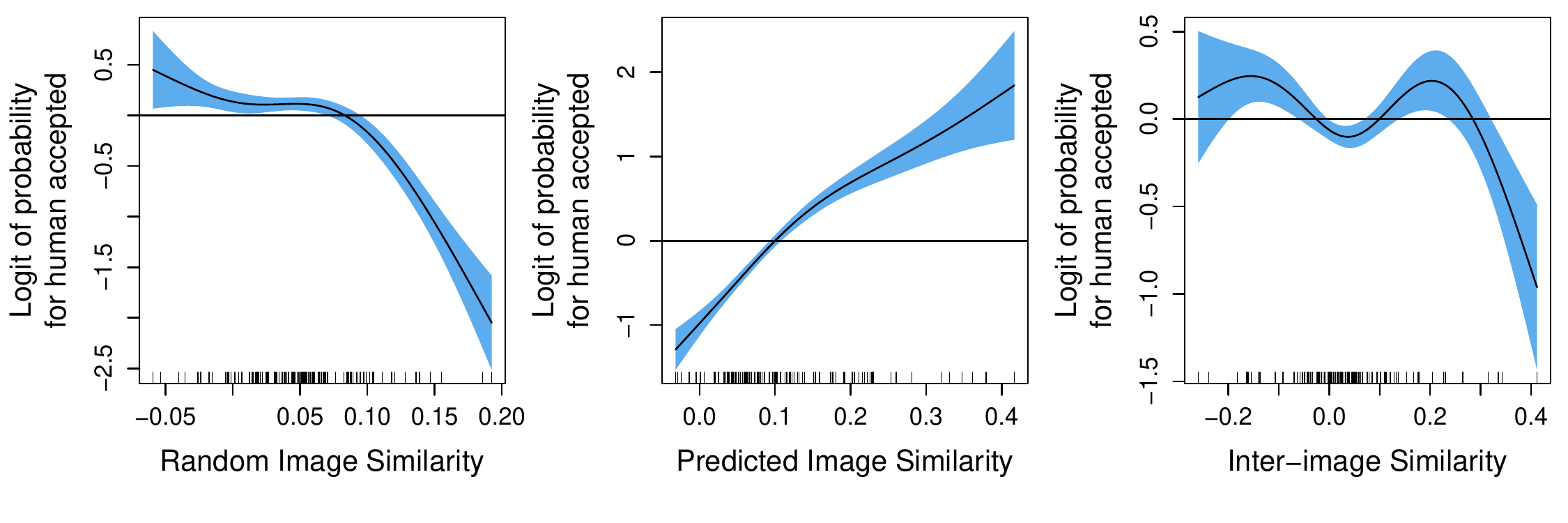}
         \caption{Prototype model}
         \label{fig:gam_proto}
     \end{subfigure}
     \hfill
     \begin{subfigure}[b]{\textwidth}
         \centering
         \includegraphics[width=\textwidth]{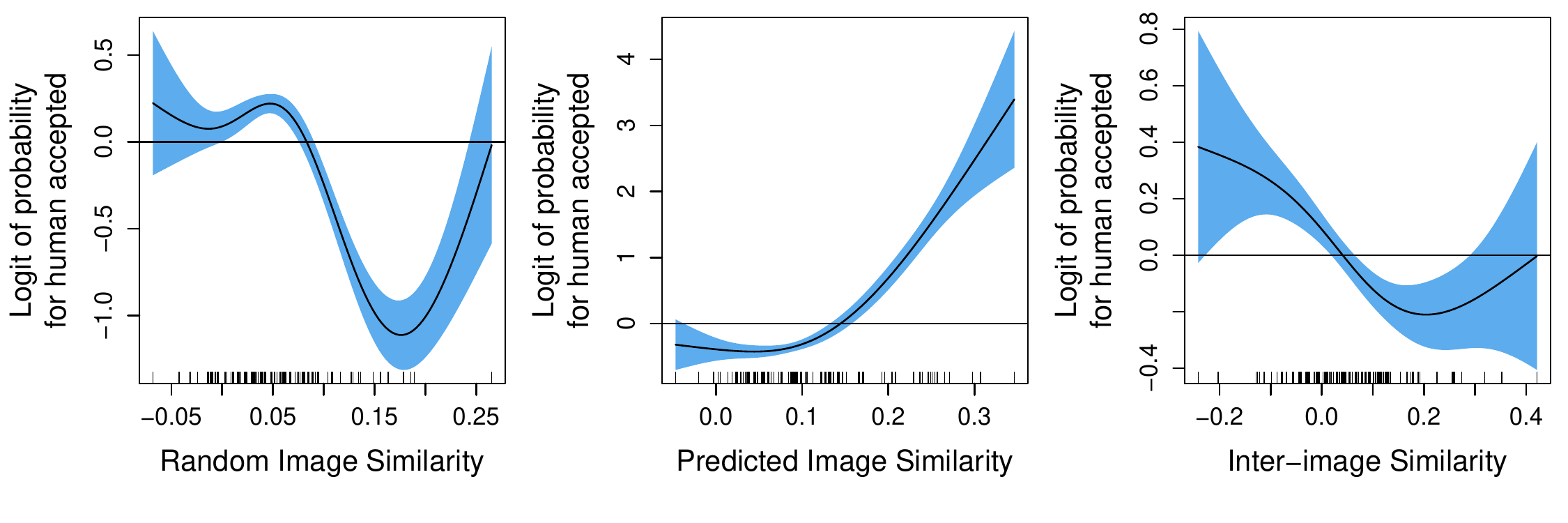}
         \caption{Exemplar model}
         \label{fig:gam_exem}
     \end{subfigure}
        \caption{Partial effects (using  thin plate regression splines) of the  predictors in GAMs for prototype and exemplar models based on grounded GloVe vectors. Plots for textual GloVe vectors as well as the Word2Vec vectors used by \citet{Gunther2020Images} can be found in the supplementary materials.}
        \label{fig:gam}
\end{figure}

The GAM also indicated (see Figure~\ref{fig:gam_proto}) that a greater \textit{Predicted Image Similarity} comes with a higher probability of selection.  This effect aligns with our hypothesis that in this task, participants are scanning images for the visible objects, basing their decision on the match between these objects and the printed word stimulus.

A greater \textit{Random Image Similarity} goes hand in hand with a lower probability of selection, but this effect is present only for higher similarity values.  Although \textit{Random Image Similarity} is in general lower than \textit{Predicted Image Similarity} (ranges (-0.06, 0.19) and (-0.03, 0.42) respectively), \textit{Predicted Image Similarity} leads to higher selection rates in the interval (0.0-0.10) whereas \textit{Random Image Similarity} does not. This suggests that the objects in the GPVM images are more tightly and consistently interconnected, so that even for low similarity values they provide consistent evidence for selection.

A non-linear effect emerged for \textit{Inter-image Similarity}. Setting aside the most extreme values of the predictor, this non-linear effect reduces to a U-shaped effect for where there is good data support.  This U-shaped effect suggests that atypical similarities (values away from the mean) induced higher ratings.  Apparently, the selection task induced an image scanning strategy that is based on whether the degree of similarity of a pair of images is remarkable and surprising. Both highly similar and very dissimilar images attract attention, resulting in more careful selection in favor of the GPVM image.

Results are subtly but informatively different for the exemplar setup. We first note that of the two factorial predictors, WordType was again supported, but Distance was not. This is in line with participant behaviour in the two setups: while they were more accurate for concrete target words than for abstract ones in both setups, the difference is much stronger in the prototype setup. Changing the kind of image --- from prototype to exemplar --- resulted in changed selection behavior.  Apparently, training the mapping on image exemplars instead of on an averaged image results in more informative images in categories with less concrete words with fewer visual neighbours. 

This has further consequences for participants' selection behavior.  Although in the prototype setup, the number of objects in the generated image (i.e. \textit{Predicted Image \#Objects}) was predictive, the number of objects in the random image (\textit{Random Image \#Objects}) was not.  However, in the exemplar experiment, the number of objects in not only the generated but also in the random image were both significant predictors of selection behavior.  For both, more objects in the image corresponded to higher selection probabilities. 

A possible explanation for the high variation in participants' selection preferences across highly concrete and abstract queries, between the exemplar and prototype models, may be attributed to the distinct training schemes utilized in each model. In the prototype setup, each noun is associated with the mean image vectors of a specific class (e.g., \textit{horse}), resulting in a feature vector conveying a typical characteristic of a given class and thereby reducing the number of feasible model outputs. Consequently, more distinct boundaries are established for concrete words, which leads to greater discriminability between GPVM images for concrete target words and random control images. In the Exemplar model, on the other hand,  each noun is linked with various image vectors that contain the target class in different contexts (other classes). For example, the word \textit{horse} may be associated with multiple images of horses in distinct settings per training sample. This results in the establishment of an association between the target noun and a diverse yet related set of classes. Although the boundaries for concrete words are not as distinct as those for the prototype model, as evidenced by a lower rating for highly concrete words, participants are more inclined to associate the target words with the different yet related set of words encountered by the model during training. Hence, in the Exemplar setup, participants are more likely to prefer the GPVM image for abstract words and concrete words in the far category. Overall, the Exemplar model might retrieve images which contain useful hints indicating its selection and encourage participants to deeply analyse the semantics of the given images with the target word, whereas the prototype model seems to be less flexible in this regard.

Further evidence for a deeper processing of the random control images is provided by the \textit{Random Image Similarity} measure, which was predictive for a larger range of values in the exemplar setup  (0, 0.20), compared to the range  (0.10, 0.20) observed for the prototype setup. For where \textit{Random Image Similarity} has dense data support, it predicted mostly a decrease in selection probability. 

Finally, a greater \textit{Inter-image Similarity} corresponded to lower selection probabilities, although for large values its effect leveled off. Whereas in the prototype setup, \textit{Inter-image Similarity} revealed that unexpected similarity boosted the selection of the generated image, in the exemplar setup, a greater similarity led to the more frequent selection of the random image.

\begin{table}[ht!]
    \begin{subtable}[h]{\textwidth}
\centering
\begin{tabular}{lrrrr}
   \hline
A. parametric coefficients & Estimate & Std. Error & t-value & p-value \\ 
  (Intercept) & -0.0351 & 0.1869 & -0.1878 & 0.8510 \\ 
  WordType=concrete & 0.5216 & 0.0851 & 6.1271 & $<$ 0.0001 \\ 
  Distance=near & 0.1866 & 0.0878 & 2.1247 & 0.0336 \\ 
  Distance=max & 0.4936 & 0.1195 & 4.1296 & $<$ 0.0001 \\ 
  Predicted Image \#Objects & 0.0928 & 0.0188 & 4.9400 & $<$ 0.0001 \\ 
  Random Image \#Objects & -0.0078 & 0.0187 & -0.4157 & 0.6776 \\ 
  WordType=concrete:Distance=near & 0.4303 & 0.1392 & 3.0913 & 0.0020 \\ 
   \hline
B. smooth terms & edf & Ref.df & F-value & p-value \\ 
  s(Random Image Similarity) & 3.5038 & 3.8671 & 111.9450 & $<$ 0.0001 \\ 
  s(Predicted Image Similarity) & 2.7066 & 3.2410 & 227.4596 & $<$ 0.0001 \\ 
  s(Inter-Image Similarity) & 3.7710 & 3.9649 & 24.0678 & 0.0001 \\ 
   \hline
\end{tabular}
       \caption{Prototype model}
       \label{tab:gam_proto}
    \end{subtable}
    \hfill
    \begin{subtable}[h]{\textwidth}
\centering
\begin{tabular}{lrrrr}
   \hline
A. parametric coefficients & Estimate & Std. Error & t-value & p-value \\ 
  (Intercept) & 0.1015 & 0.1689 & 0.6009 & 0.5479 \\ 
  WordType=concrete & 0.3432 & 0.0906 & 3.7898 & 0.0002 \\ 
  Distance=near & 0.1636 & 0.0931 & 1.7574 & 0.0789 \\ 
  Distance=max & -0.1399 & 0.1139 & -1.2286 & 0.2192 \\ 
  Predicted Image \#Objects & 0.0544 & 0.0202 & 2.6973 & 0.0070 \\ 
  Random Image \#Objects & 0.0704 & 0.0177 & 3.9813 & 0.0001 \\ 
  WordTypeconcrete:distance\_near1 & -0.1100 & 0.1399 & -0.7859 & 0.4319 \\ 
   \hline
B. smooth terms & edf & Ref.df & F-value & p-value \\ 
  s(Random Image Similarity) & 3.8844 & 3.9923 & 132.6433 & $<$ 0.0001 \\ 
  s(Predicted Image Similarity) & 3.2650 & 3.6982 & 184.3976 & $<$ 0.0001 \\ 
  s(Inter-Image Similarity) & 2.7887 & 3.3389 & 25.7459 & $<$ 0.0001 \\ 
   \hline
\end{tabular}
        \caption{Exemplar model}
        \label{tab:gam_exem}
     \end{subtable}
     \caption{GAM summary tables for prototype and exemplar models for grounded GloVe. Summary tables for textual/grounded W2V models can be found in the supplementary materials.}
     \label{tab:gam}
\end{table}

\subsection{Q2: Is participants’ behaviour best accounted for by purely textual or multimodal word embeddings?}

The previous subsection discussed results for both purely textual and grounded embeddings without further discussing any differences in their performance. However, our second question is whether using visually grounded embeddings instead of purely textual embeddings will improve prediction accuracy.

Similar to the previous subsection, we first focus on the Max approach selecting the GPVM image or the random control image for each target word. 
If participants generated visual images for the target word, then the grounded vectors developed by \citeauthor{shahmohammadi2022language} are expected to provide enhanced prediction accuracy for participants' behavior. 
Table~\ref{tab:virtual} shows that using grounded embeddings 
results in lower $\Delta$ values for both grounded GloVe and grounded Word2Vec embeddings in the exemplar model, and for grounded GloVe embeddings in the prototype model, suggesting that the grounded embeddings model participants' preference somewhat better compared to textual embeddings. We conducted a sign test for both setups, comparing the average accuracy in each WordType/Distance category between the textual and grounded version of each embedding type, thus resulting in 10 comparisons overall.  We counted a comparison as a  ``success'' if the average accuracy of the grounded embeddings was closer to participants' performance than the textual one.  Interestingly, the sign test clarified that actually the grounded embeddings were not significantly better than their purely textual counterparts, neither in the prototype nor in the exemplar setup. 

Moving on to predicting participants' behaviour directly, we find that for the prototype setup (see Table~\ref{tab:res_proto}), using visually grounded vectors improves mean accuracy by 2\% for  GloVe and 1\% for Word2Vec (compare Max: GloVe with  Max: ZSG-GloVe, and Max: W2V with Max:ZSG-W2V).   However, for the exemplar setup (see Table~\ref{tab:res_exem}), accuracy decreased by 4\% for ZSG-Glove and by 1\% for ZSG-W2V.

Turning to the results of the GAMs (rows denoted ``GAM'', column ``Mean'' in Table~\ref{tab:marco}), we observe that the GAMs for the prototype models show better accuracy (by 1-2 percentage points) when based on grounded rather than on purely textual embeddings. In the exemplar setup, they do not show better results in predicting human responses. In terms of AIC, the GAMs for both prototype and exemplar models show a better model fit when based on grounded GloVe embeddings than on textual ones (by 80.4 and 86.74 AIC points respectively) and on grounded Word2Vec embeddings only in the prototype setup (by 123.1 AIC points; in the exemplar setup the difference was -3.1 points), again compared to textual ones. 

In summary, in terms of numerical differences, the evidence of multimodal embeddings improving prediction for participants' behaviour is mixed. For GPVM images in the prototype setup, both the Max and GAM evaluation methods show a slight advantage for grounded embeddings. For exemplar images, no such advantage is visible in neither the Max nor the GAM evaluation. We again ran a sign test for the prototype and exemplar setups respectively in the same way as in the previous section (20 comparisons per setup, this time a comparison was a ``success'' if the grounded version of the embeddings showed a larger accuracy). The difference was again not significant in both setups. 

The reason that the grounded embeddings by \citeauthor{shahmohammadi2022language} are on average, numerically, somewhat less effective for the exemplar model is likely to be that grounded embeddings cluster by semantic similarity rather than by semantic relatedness \citep{shahmohammadi2021learning, shahmohammadi2022language}.  Since in the exemplar model, due to the way in which images are processed, relatedness plays a much stronger role than in the prototype approach (see Section~\ref{sec:q1}), the grounded embeddings are less effective for the experimental data obtained from the exemplar-based set-up.

Considered jointly, these results lead us to conclude that participants' behaviour appears to be equally well accounted for by purely textual vectors and multimodal vectors.  This result raises doubts about participants actually generating visual images of the target words.

\subsection{Q3: Does the indirect grounding of abstract words afford a better understanding of the experimental results reported by GPVM?} 

Is visual grounding beneficial not only for concrete words but also for abstract words? On the basis of a series of human-annotated semantic similarity datasets, \citeauthor{shahmohammadi2022language} argued that indeed abstract words do benefit from indirect visual grounding.  Can the same conclusion be drawn for the data of GPVM? 

Analogously to the previous two subsections we again first consider the proportions of the Max approach selecting the GPVM image over the random control image, this time broken down for each of the combinations of WordType and Distance. Considering Table~\ref{tab:virtual_proto}, in the prototype experiment, participants' scores are close to random for the abstract far condition, they are somewhat higher for the abstract near and concrete far conditions, and they are highest for the concrete near and concrete max conditions. Focusing on the best model, Max: ZSG-GloVe, we find a very similar pattern. Comparing this model to the predictions of Max: GloVe, we find that grounding moves the predictions close to human performance in all conditions except the concrete max. The most notable difference here can be found in the abstract far category, where Max: GloVe selects the GPVM image far more often than human participants (see also Figure~\ref{fig:acc_proto}).

Next, consider Table~\ref{tab:virtual_exem}, which concerns participants' selection preferences for the GPVM images in the exemplar setup.  Compared to their performance in the prototype setup, participants' accuracy scores are down considerably for Concrete Near and Concrete Max, and up considerably for Abstract Far and Abstract Near.  Performance for Abstract Far words clearly lags behind performance for the other four subsets of words.  The GPVM images in the exemplar setup elicited flatter scores, consistent with our conclusion in the preceding sections that in this setup participants scan the control images more carefully. Both the Max:ZSG-GloVe and Max:ZSG-W2V models perform reasonably similar to the participants' preferences, but there are conditions where textual embeddings capture their preferences better. However, it should be noted that all models are well within the range of participants' performance (Figure~\ref{fig:acc_exem}).

Next, we turn to predict participants' selection behaviour directly. First consider Table~\ref{tab:res_proto}, which concerns the prototype setup. Focusing on the best model, GAM:ZSG-GloVe, we find that prediction accuracy is clearly higher for Concrete Near and Concrete Max compared to the Abstract and Concrete Far conditions.  Comparing this model with GAM:GloVe, we see that  visual grounding improves accuracies for 3 of the five subsets: Concrete Far, Concrete Max, and Abstract Far.  There is one subset where grounding leads to lower scores,  Abstract Near, and one where grounding does not change performance, Concrete Near. Averaging over both subsets of abstract words, it seems that there is a modest advantage overall for the visual grounding of abstract words.

Table~\ref{tab:res_exem} concerns the exemplar setup. The GAM:ZSG-GloVe model performs reasonably similar to the participants. Compared to GAM:GloVe, for the abstract words, the model shows an improvement of 4\% in the Abstract Far category and a reduction in accuracy of 4\% for the Abstract Near category.

In summary, it appears that visual grounding aligns more closely with participants' selection behavior in the prototype setup, but its effect is somewhat mixed for abstract words in the exemplar setup. A potential explanation for this finding, same as in the previous section, is that visual grounding tends to create clusters of similar words rather than clusters of related words \citep{shahmohammadi2021learning, shahmohammadi2022language}. Given that the exemplar model establishes an association between the target nouns and diverse yet related concepts, its behavior for abstract nouns may be better explained by textual embeddings. Therefore, shifting the focus toward similarity appears to benefit highly concrete words but has a negative impact on modeling abstract words.

\section{Discussion and Conclusion}\label{sec:discussion}

We started this investigation with three questions: first, can we predict the behaviour of participants in the experiments reported by \citet{Gunther2020Images} without assuming that they generated mental images? Second, is participant behaviour predicted better by visually grounded or purely textual word embeddings? And third, how does the visual grounding process affect performance on abstract words?

Regarding the first question, we found that 
{ 
an approach taking into account the objects present in the presented images is able to predict participants' behaviour quite well. The covariates that we derived from the embeddings for the objects in the images, random image similarity, predicted image similarity, and inter-image similarity, all helped improve the logistic GAMs that we fitted to predict participants' choice behavior.  This finding dovetails well with the eye-tracking literature on image scanning: typically, images illicit multiple fixations, reflecting attention being directed to different parts of images and different objects in images \citep{cronin2020eye}. From these findings, we infer that participants probably based their decisions on comparisons in semantic space, and not only in visual space.
Our experiments suggest that it is unlikely that participants really generated mental images for the words presented to them.  Although eye-tracking  experiments suggest that participants can get the gist of an image within a time span of 40ms \citep[see][for a review]{castelhano2008eye}, understanding images usually requires a series of fixations.  This finding does not fit well with the assumption made by GPVM that the images presented to the participants in the experiments of GPVM were processed holistically and were compared with an equally holistic image projected from the target word.   Furthermore, it is well-established that our perception of the world is shaped by the limitations of our sensory organs and the constraints imposed by the cultures we live in \citep[see, e.g.,][]{kant1999critique,hoffman2019case}. The way in which \citeauthor{shahmohammadi2022language} implement visual grounding --- constraining the extent to which vision can change embeddings from human texts --- does justice, however crude, to this insight.
}

{ Our conclusions are also} in line with previous findings on mental imagery: for example, \citet{Louwerse2011AWords} argue that modality-information (such as visual information) is to some extent already included in linguistic information and that only for more precise information embodied simulation is required (thus arguing that both linguistic and embodied processes contribute to conceptual processing). According to their study, linguistic processes account for early processing (short reaction times) and embodied ones for later processing (longer reaction times). Our results also dovetail well with the views on grounding proposed by \citet{Zwaan2005EmbodiedComprehension} and \citet{Barsalou1999PerceptualSystems} mentioned in the introduction. 

As to question 2, we found that our models for predicting participants' performance are slightly, though not statistically significant, improved by using grounded embeddings compared to purely textual embeddings for GPVM's 
prototype setup.


While there was a slight numerical improvement in the prototype setup, \citeauthor{shahmohammadi2022language}'s grounded embeddings were not able to improve on the textual baseline in the exemplar model. Our interpretation of this result is that the images predicted by GPVM 
in the exemplar setup are
driven more by semantic relatedness than by semantic similarity (by virtue of how the model is trained), and as the visual grounding method of \citeauthor{shahmohammadi2022language} enhances semantic similarity rather than semantic relatedness, it is less effective for the experimental data of the exemplar model.  

{
GPVM argued that their exemplar model picked up more ``idiosyncratic information'', leading to a loss of predictivity of concreteness and number of visual neighbors.  The above comparison of the performance of textual and grounded embeddings suggests that the exemplar model is not picking up just noise (idiosyncatic information), but rather that it is more influenced by semantic relatedness, mediated by the objects that co-occur with the  objects that are actually targeted in the images used to train the models (e.g., a doctor co-occurring in an image selected to depict a nurse). The employed visually grounded vectors, by their design zoom in on semantic similarity, whereas standard textual vectors are somewhat more sensitive to semantic relatedness.  These considerations lead us to conclude that in the experiments of GPVM, subjects' decisions were guided by both semantic similarity and semantic relatedness, and that the way in which images were selected (prototype vs. exemplar) influenced the relative importance of similarity and relatedness in participants' decision making.  
}

Here, it is important to stress the differences between the grounding model proposed by GPVM 
and the one by \citeauthor{shahmohammadi2022language}. 
GPVM 
posit a simple linear mapping from textual to visual embeddings. This model works as a proof of concept that a connection between language and vision can be drawn. However, it is not necessarily the best way to combine the two modalities. \citeauthor{shahmohammadi2022language} showed that the embeddings generated by such a simple linear mapping perform much worse at predicting human ratings in a number of semantic similarity and relatedness datasets. This indicates that it is not enough to show that language and vision \textit{can} be linked: the real challenge is to understand \textit{how} humans \textit{combine} information from both modalities in order to form meaning representations and make similarity judgments. 

Regarding question 3, we observed that grounding tends to yield better predictions of human judgments than textual embeddings for abstract words at least in the prototype setup. This finding, together with previous results suggesting that grounding improves performance on similarity/relatedness judgment tasks even for abstract words \citep{shahmohammadi2022language}, begs the question of why a grounding model trained only on concrete words for which images were available in COCO improves performance also for abstract words. Our interpretation of these results is that by improving the relative position of concrete words in semantic space, the representations of abstract words are also improved. In other words, the visual alignment trained on concrete words also benefits abstract words by transferring them into a more precise  semantic space. This view is in line with conceptual metaphor theory which posits that abstract words are understood in terms of concrete words \citep[e.g.][]{lakoff1980metaphorical}.

With respect to how abstract and concrete words are learned, \citet{vigliocco2018learning} conclude that only by the age of 10 children have sufficient experience with their language to be able to start making use of distributional semantics for learning abstract words.  They argue that children are more likely to be using the strong association between abstractness and emotional valence for learning abstract words.  Images come with emotional values that are visible in the EEG even under scrambling \citep{rozenkrants2008affective}.  By visually grounding abstract words, it is possible that the embeddings of abstract words are not only more precisely profiled with respect to concrete words, but also that abstract words are better grounded with respect to their emotional loadings. 
Empirical studies have provided extensive support for the significant role of emotion in human cognition, especially in abstract concepts \citep[see][for a review]{Dolan:2002}. It has been shown that the addition of emotional representations to textual embeddings improves classification tasks on datasets that contain mostly abstract words \citep{Rotaru:Vigliocco:2020}.
It is also possible that once visually grounded embeddings are used, instead of purely textual vectors, children may turn out to be sensitive to distributional aspects of their language at an earlier age than reported by \citet{vigliocco2018learning}.

We finish with three comments: Firstly, with regard to other visually grounded models: Despite the tremendous amount of work on the visual grounding of textual embeddings, the common belief holds that grounding words visually is beneficial for words with concrete meaning and has an adverse effect on abstract words \citep{pezzelle2021word,kiros-etal-2018-illustrative,kiela-etal-2018-learning,park2017computational}. Some new studies further suggest that visual grounding of current sentence-level contextualized textual models does not add extra knowledge for downstream NLP tasks \citep{Yun2021DoesVP,tan-bansal-2020-vokenization}. However, the common core idea among many previous grounded approaches is fusing vision and language into a single modality. That is, image vectors and word/sentence vectors are either 1. mapped to a common semantic space where similar visual and textual concepts are forced to have similar representations, most often using non-linear transformation, or 2. word vectors of concrete words are replaced by image vectors during the training process. 

\citet{shahmohammadi2022language} on the other hand showed that allowing the complete fusion of both modalities,  while benefiting the concrete words, is detrimental to modeling abstract words. They argue that language benefits from vision the most once it is guided by perceptual knowledge as opposed to being merged with it. Using this idea,  \citeauthor{shahmohammadi2022language} showed that visual grounding is highly beneficial for modeling abstract words and further boosts the performance on downstream NLP tasks when limited training is available. Even though their model learns a linear alignment based on a limited number of captions describing concrete scenes, it was used to indirectly generate grounded representations for unseen abstract words. This is in line with the indirect grounding perspective that implies the direct grounding of concrete words and indirect grounding of abstract words via language \citep{howell2005model,louwerse2011symbol}. The indirect grounding theory of abstract words has been recently shown effective at predicting abstract concepts using distributional semantic models \citep{utsumi2022test}. Indirect grounding, therefore, seems to be a plausible cognitive mechanism for grounding abstract words.

Secondly, as a final conclusion regarding GPVM, 
we note that while their experiment clearly highlights that a simple linear mapping is able to predict images that are chosen by participants above chance level, our work highlights that the conclusions which can be drawn from this study are far from clear. Firstly, we know from previous research \citep{shahmohammadi2022language} that a simple linear mapping as used by GPVM 
is not a good grounding model: vectors that are grounded in such a way perform much worse than purely textual embeddings on word similarity and relatedness datasets. Secondly, we are able to demonstrate that the task which GPVM
used to evaluate their embeddings can be solved to a large extent without taking into account grounded meaning representations. Thus, it is unclear how much the experiment actually taps into aspects of grounded meaning representations. In this study, we used grounded embeddings which have been shown to improve upon purely textual embeddings in previous work \citep{shahmohammadi2022language}. We found a modest numerical, though not statistically significant, improvement of these embeddings over purely textual embeddings in the prototype setup and no major improvement in the exemplar setup. This suggests to us that if at all, the effects of grounding are more pronounced in the prototype setup.  Thus, the method for grounding used in 
GPVM 
may not provide the optimal method for generating grounded word representations that can be utilized in psycholinguistics tasks, and the extent to which their experiment is well-suited for detecting effects of grounded meaning representations in human cognition remains somewhat uncertain. 

Thirdly, what can we ultimately conclude from this about how humans represent meanings? On the one hand, the present paper together with the results from \citeauthor{shahmohammadi2022language} support the conclusion that meaning representations are not based on language alone, but also include information based on vision\footnote{Note that we restricted ourselves to visual information here. It is very likely that other multimodal information such as auditory and olfactory information also plays a role, see \citet{kiela2015grounding, Kiela2015Multi-Perception}}. Moreover, both \citeauthor{shahmohammadi2022language}, as well as the present study, show that this applies not only to concrete words, where such an effect may be expected but also to abstract ones. On the other hand, however, \citeauthor{shahmohammadi2022language} found that if textual information is overwhelmed by image information, resulting embeddings as predictors of human similarity ratings suffer. This dovetails well with many previous studies which reported that representations based on textual information alone can predict behavioural data very successfully \citep[e.g.][]{Mandera2017ExplainingValidation, Westbury2014YouJudgments, Westbury2019WrigglyFunny}. These results underline that human meaning representations are largely based on the experience humans have with the world through language. 

\section{Acknowledgements}

This work has been { funded by the Deutsche Forschungsgemeinschaft (DFG, German Research Foundation) under Germany’s Excellence Strategy – EXC number 2064/1 – Project number 390727645} as well as by the German Federal Ministry of Education and Research (BMBF): Tübingen AI Center, FKZ: 01IS18039A, and the European research council,  project WIDE-742545. The authors thank the International Max Planck Research School for Intelligent Systems (IMPRS-IS) for supporting Hassan Shahmohammadi. 

\vfill

\bibliography{sample}

\end{document}